\documentclass[lettersize,journal]{IEEEtran}
\usepackage[caption=false,font=normalsize,labelfont=sf,textfont=sf]{subfig}

\usepackage{amsmath,amssymb}
\usepackage{fix-cm}
\usepackage{array,booktabs,multirow}
\usepackage{graphicx}
\usepackage{textcomp}
\usepackage{helvet}
\usepackage{courier}
\usepackage[table]{xcolor}
\usepackage[small]{caption}
\usepackage{url}
\usepackage{stfloats}
\usepackage[hidelinks]{hyperref}
\hypersetup{
  pdftitle={NatureFusion: Rethinking Infrared and Visible Image Fusion as Natural Complementary Enhancement},
  pdfauthor={Jianfeng Ding, Linfeng Tang, Hao Zhang, Han Xu, Xunpeng Yi, Yibing Zhang, Meiqi Gong, Zhongyuan Wang, Jiayi Ma},
  pdfkeywords={Natural complementary enhancement, generative prior transfer, infrared and visible image fusion, task-driven fusion}
}

\definecolor{bestred}{RGB}{248,231,233}
\definecolor{secondblue}{RGB}{231,240,251}
\begin{document}

\title{NatureFusion: Rethinking Infrared and Visible Image Fusion as Natural Complementary Enhancement}

\author{Jianfeng~Ding, Linfeng~Tang, Hao~Zhang, Han~Xu,\\
        Xunpeng~Yi, Yibing~Zhang, Meiqi~Gong, Zhongyuan~Wang, and~Jiayi~Ma\thanks{This work was supported by the National Natural Science Foundation of China (62676289). \textit{(Corresponding authors: Zhongyuan Wang and Jiayi Ma.)}}\IEEEcompsocitemizethanks{\IEEEcompsocthanksitem Jianfeng Ding and Zhongyuan Wang are with the National Engineering Research Center for Multimedia Software (NERCMS), School of Computer Science, Wuhan University, Wuhan 430072, China (e-mail: jianfeng0369@gmail.com, wzy\_hope@163.com).
\IEEEcompsocthanksitem Linfeng Tang is with the School of Robotics, Wuhan University, Wuhan 430072, China (e-mail: linfeng0419@gmail.com).
\IEEEcompsocthanksitem Hao Zhang, Xunpeng Yi, Yibing Zhang, Meiqi Gong, and Jiayi Ma are with the Electronic Information School, Wuhan University, Wuhan 430072, China (e-mail: zhpersonalbox@gmail.com, yixunpeng@whu.edu.cn, zhangyibing@whu.edu.cn, meiqigong@whu.edu.cn, jyma2010@gmail.com).
\IEEEcompsocthanksitem Han Xu is with the School of Automation, Southeast University, Nanjing 210096, China (e-mail: xu\_han@seu.edu.cn).}}

\maketitle

\begin{abstract}
In the absence of ground truth, most existing infrared and visible image fusion methods pursue fidelity to the source images and directly superimpose two physically distinct modalities. As a result, salient infrared targets appear as bright patches of thermal radiation pasted onto the scene. These patches obscure the original color and texture of the targets, so the fused images look unnatural to human observers and deviate from the data distribution learned by vision models pretrained on visible images. Therefore, we redefine infrared and visible image fusion as \textbf{natural complementary enhancement} rather than source fidelity, and propose \textbf{NatureFusion}, which exploits cross-modal complementary cues to infer a clear and natural visible-style image from low-quality visible and infrared image pairs. Specifically, to address the lack of ground truth, we employ a generative foundation model as the teacher to directly generate visually natural fused images for supervision. Furthermore, to fully distill the teacher's generative capability while reducing computational cost, we fine-tune a pretrained generative model with conditional flow matching through a light-weight fusion adapter. Last but not least, to further improve downstream performance without sacrificing natural appearance, we freeze the fusion adapter and introduce a detector-supervised task adapter that strengthens the complementary contribution of infrared. Extensive comparisons with seven multi-modal fusion and seven single-modality enhancement methods demonstrate that NatureFusion generates visually natural fused images with the best perceptual quality. Detection and segmentation models pretrained on visible images achieve substantially higher accuracy on our fused images than on those of any competing method.
\end{abstract}

\renewcommand{\topfraction}{0.7}

\section{Introduction}
\label{sec:introduction}

\begin{figure}[!t]
    \centering
    \includegraphics[width=\columnwidth]{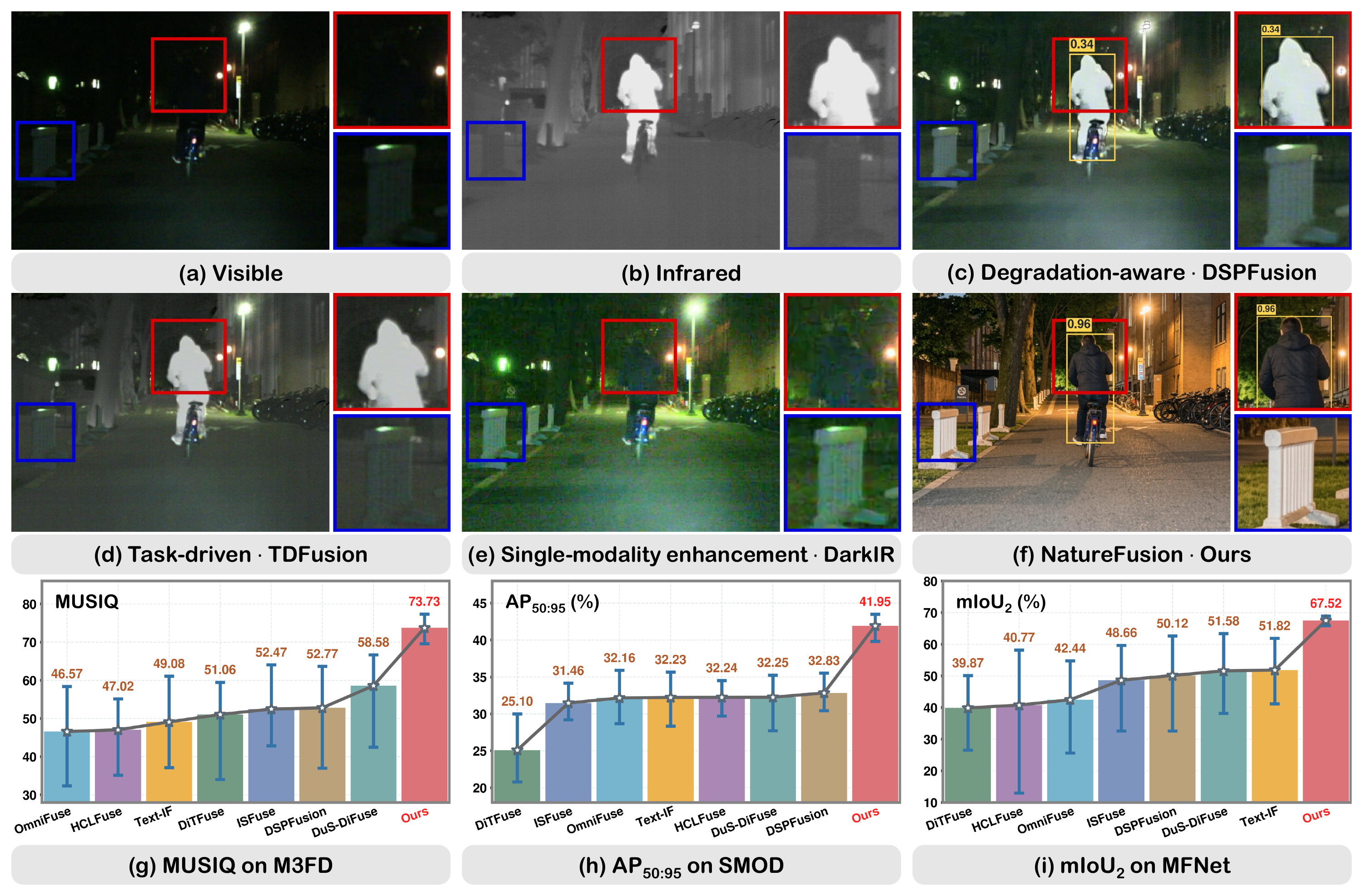}
    \caption{Comparison with multi-modal fusion and single-modality enhancement methods. Panels (a) and (b) show the visible and infrared inputs, (c) to (e) show the results of degradation-aware fusion, task-driven fusion, and single-modality enhancement methods, and (f) shows the result of NatureFusion. Red and blue boxes mark enlarged person and background regions, and gold boxes show person detections of a frozen RetinaNet with confidence scores of at least 0.25. (g) to (i) Comparison with seven multi-modal fusion methods in MUSIQ on M3FD, person $\mathrm{AP}_{50:95}$ on SMOD, and $\mathrm{mIoU}_2$ over Car and Person on MFNet. Detection and segmentation each use seven frozen models pretrained on visible images. Bars show means, and whiskers show ranges across images and models.}
    \label{fig:motivation}
\end{figure}

Visible images capture the color and texture of a scene, but targets such as pedestrians are often difficult to recognize in poorly lit regions. Infrared images record thermal radiation from the scene and can highlight the locations of these targets, yet they lack color and texture details, as shown in Fig.~\ref{fig:motivation}(a) and (b). Therefore, infrared and visible image fusion aims to combine the complementary information of the two modalities into a fused image that is better suited to observation and analysis, benefiting both human observers in applications such as in-vehicle night vision and vision models in downstream tasks such as object detection.

However, in the absence of ground truth specifying what an ideal fused image should look like, most existing infrared and visible image fusion methods follow a source-fidelity paradigm, which uses unsupervised intensity, structure, and gradient losses to keep the fused image faithful to the source images. Notably, this paradigm faces a critical challenge, namely that the two modalities are physically distinct. Infrared intensity reflects the thermal radiation of objects rather than their appearance under visible light, so a pedestrian in dark clothing can appear bright in the infrared image but dark in the visible image. When the fused image is forced to remain faithful to both modalities, such targets are highlighted as bright patches of thermal radiation that obscure their original color and texture, making the fused image unsuitable for both human observers and vision models, as shown in Fig.~\ref{fig:motivation}(c) and (d). For human observers, these patches make the fused image look unnatural, especially in low-light and other degraded scenes, where weak visible signals leave the result dominated by thermal patterns. To cope with such scenes, degradation-aware methods incorporate image enhancement, restoration, or generative priors into fusion~\cite{tang2023divfusion,yi2024textif,zhao2023ddfm,zhang2025omnifuse}. Although they improve scene illumination and texture, they still pursue source fidelity, leaving the thermal patches on targets intact. For vision models, these patches push the fused image away from the data distribution of visible images, degrading the predictions of detection and segmentation models pretrained on visible images. To address this issue, task-driven methods introduce detection or segmentation losses and train the fusion network jointly with the task network~\cite{tang2022seafusion,bai2025tdfusion}. Nevertheless, these methods usually still build on source-fidelity losses and cannot fundamentally improve the naturalness of the fused images. As a result, their performance gains largely rely on fine-tuning the task network on the fused results and can hardly transfer to existing models pretrained on visible images. Moreover, the complementary value of infrared and visible image fusion lies particularly in targets that are hard to recognize in the visible image but salient in the infrared image. Brightening the visible image alone, as single-modality low-light enhancement does~\cite{yi2026ll}, can hardly recover these targets, as shown in Fig.~\ref{fig:motivation}(e). In other words, making these targets discernible depends on the location and contour cues provided by the infrared image. However, task losses mainly constrain the final predictions and impose no explicit constraint on the contribution of each modality, so the optimization may rely on discriminative features already present in the visible image and fail to fully exploit the complementary information of infrared.

To address these problems, we redefine infrared and visible image fusion as \emph{natural complementary enhancement}. The core idea is to exploit cross-modal complementary cues to generate a clear and natural visible-style image, rather than to preserve and superimpose the information of the two modalities directly. Specifically, the visible image determines how targets should look, providing scene color and texture to constrain the appearance of the fused result. The infrared image indicates where targets are and provides their contour cues, guiding the enhancement of targets that are hard to recognize in the visible image. With the prior learned by a generative foundation model, infrared targets are no longer directly superimposed as bright patches of thermal radiation. Instead, they are rendered with visible colors and textures consistent with the surrounding scene and become clearly recognizable, as shown in Fig.~\ref{fig:motivation}(f). As a result, the fused image not only has a natural appearance that agrees with human visual perception but also lies closer to the data distribution of visible images, providing clearer target information for downstream detection and segmentation models that are pretrained on visible images.

To realize natural complementary enhancement, we propose NatureFusion, a generative image fusion framework that uses two decoupled adapters for natural fusion and task adaptation, as shown in Fig.~\ref{fig:pipeline}. More specifically, to address the lack of fusion ground truth, we introduce a generative foundation model as a teacher, which generates fused images with both a natural appearance and target discriminability under the joint guidance of the visible image and infrared target cues. Since this teacher is costly to query, we distill its generative capability into a pretrained generative model by training only a light-weight fusion adapter with conditional flow matching. Benefiting from the natural-image prior of the pretrained model, the fusion adapter learns effectively from limited demonstrations and enables natural complementary enhancement from infrared and visible image pairs alone. Building on these natural results, which already lie close to the data distribution of visible images, we further train the task adapter to improve downstream performance. We keep the generative model and the fusion adapter frozen and supervise the task adapter with a frozen detector pretrained on visible images. In this way, task optimization adapts the fused results to existing vision models while leaving the learned natural mapping unchanged, and the resulting gains carry over to other models pretrained on visible images. As noted above, the detection loss constrains only the final predictions, so the task adapter may still rely on discriminative features already present in the visible image. To prevent this, we use the difference between predictions with and without infrared to explicitly strengthen infrared's complementary contribution.

Our main contributions are as follows.
\begin{itemize}
\item We propose natural complementary enhancement, a new fusion paradigm that converts infrared target cues into a scene-consistent visible appearance rather than superimposing them, serving both humans and vision models.
\item We propose a fusion learning method guided by a generative foundation model, which constructs fusion supervision from multi-modal inputs and trains a fusion adapter with conditional flow matching for natural fusion.
\item We design a decoupled task adaptation method that improves downstream task performance through detection supervision and the enhancement of complementary information, while preserving the natural fusion capability.
\item Extensive experiments demonstrate that NatureFusion generates natural and visually pleasing results and substantially improves the performance of various detection and segmentation models pretrained on visible images.
\end{itemize}

\raggedbottom
\section{Related Work}
\label{sec:related_work}

\subsection{Degradation-Aware Multi-Modal Image Fusion}
\label{subsec:rw_degradation}

Under adverse imaging conditions, degradations such as low illumination and noise propagate from the source images into the fused results. To address this issue, degradation-aware fusion incorporates degradation handling into cross-modal information integration. Early methods start by adjusting the contribution of each modality. For example, PIAFusion~\cite{tang2022piafusion} uses the illumination probability to adjust the weights of the two modalities in the fusion loss. Later, DIVFusion~\cite{tang2023divfusion} removes the illumination degradation in the visible image before fusion, while Text-IF~\cite{yi2024textif} unifies restoration and fusion in a single model under the guidance of textual semantics. Another line of work introduces generative priors. DDFM~\cite{zhao2023ddfm} performs fusion using the image prior learned by a pretrained diffusion model. OmniFuse~\cite{zhang2025omnifuse} and DuS-DiFuse~\cite{cao2026dusdifuse} further extend this idea to degraded scenes by jointly modeling degradation removal and fusion in the latent space. These methods have progressively improved fusion quality under degradation. However, they still follow the source-fidelity paradigm, so infrared targets in their fused results appear as bright patches of thermal radiation. In contrast, NatureFusion uses natural fused images generated by a generative foundation model as supervision and renders infrared targets with a visible appearance consistent with the surrounding scene.

\flushbottom
\subsection{Task-Driven Multi-Modal Image Fusion}
\label{subsec:rw_task}

\looseness=1
Task-driven fusion introduces supervision from downstream vision tasks such as detection and segmentation to make fused images more suitable for machine perception. SeAFusion~\cite{tang2022seafusion} uses a semantic segmentation loss to guide the training of the fusion network, and SegMiF~\cite{liu2023segmif} links fusion and segmentation features through hierarchical interactive attention. TDFusion~\cite{bai2025tdfusion} learns a fusion loss under task supervision, while DCEvo~\cite{liu2025dcevo} balances the optimization of fusion and detection through evolutionary learning. These methods have improved downstream performance. However, they usually still build on source-fidelity losses and thus produce unnatural fused results, and their task losses impose no explicit constraint on the infrared contribution. In contrast, NatureFusion performs task adaptation on top of natural fused results and explicitly strengthens the \mbox{complementary contribution of infrared.}

\subsection{Single-Modality Image Enhancement}
\label{subsec:rw_lowlight}

\looseness=1
Unlike the two fusion paradigms above, natural complementary enhancement no longer pursues source fidelity and instead generates a natural visible-style image. Single-modality image enhancement likewise aims to improve the natural appearance of visible images. This subsection therefore reviews such methods, with a focus on low-light image enhancement. Based on Retinex theory, Retinexformer~\cite{cai2023retinexformer} estimates illumination to brighten the image and removes the noise in dark regions as well as the artifacts introduced by brightening. HVI-CIDNet~\cite{yan2025hvi} proposes the HVI color space, in which brightness and color are decoupled and enhanced separately, and DarkIR~\cite{feijoo2025darkir} performs robust low-light image restoration. Generative priors have also been introduced into this field. DiffLL~\cite{jiang2023diffll} and Diff-Retinex++~\cite{yi2025diffretinexpp} enhance images with diffusion models, and QuadPrior~\cite{wang2024quadprior} and MR. Illuminate~\cite{cho2026mrilluminate} achieve zero-reference and zero-shot enhancement with physical and diffusion priors, respectively. These methods rely on visible observations and image priors to improve visual quality. However, they cannot reliably recover targets that are barely recognizable in the visible image. Image priors can suggest plausible colors and textures, but provide no additional observation of a target's presence or location in the current scene. In contrast, NatureFusion obtains the locations and contours of such targets from the infrared image, making them \mbox{clearly recognizable in the visible-style result.}

\begin{figure*}[!t]
    \centering
    \includegraphics[width=\textwidth]{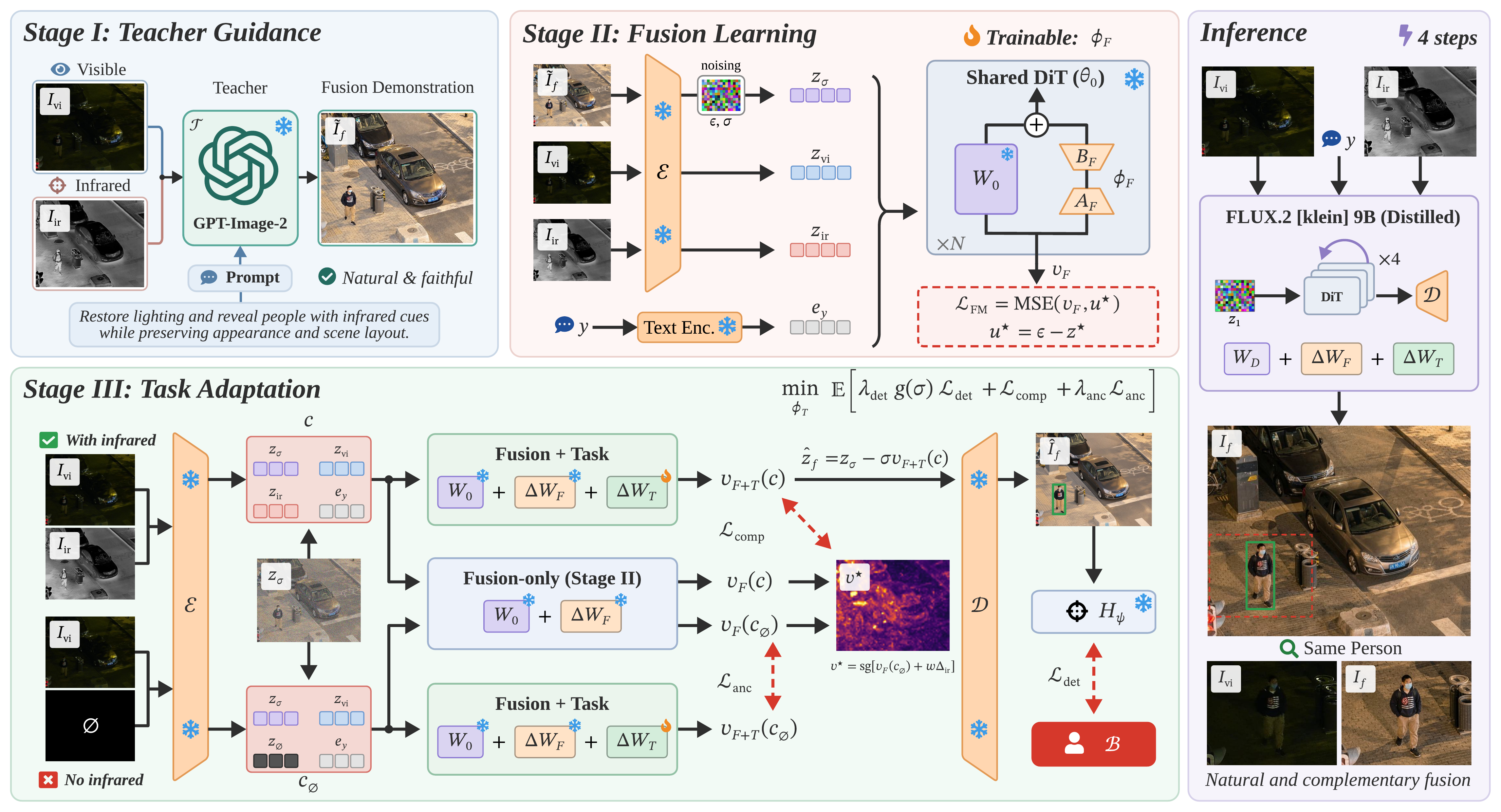}
    \caption{Training and inference pipeline of NatureFusion. Flames and snowflakes denote trainable and frozen parameters, respectively.}
    \label{fig:pipeline}
\end{figure*}

\section{Method}
\label{sec:method}

\subsection{Problem Formulation}
\label{subsec:formulation}

Given registered visible and infrared images of the same scene, $I_{\mathrm{vi}}\in\mathbb{R}^{H\times W\times 3}$ and $I_{\mathrm{ir}}\in\mathbb{R}^{H\times W}$, NatureFusion generates an RGB fused image $I_f\in\mathbb{R}^{H\times W\times 3}$. Under the proposed natural complementary enhancement principle, $I_f$ should resemble a visible image captured under more favorable imaging conditions. Infrared reveals the locations and contours of targets that are difficult to discern in the visible image, while the visible image provides scene-consistent color and illumination cues to render these targets with natural visible appearance. Learning this mapping presents two challenges. First, such an ideal $I_f$ cannot be directly captured, leaving no ground truth for supervision. Second, downstream task adaptation may affect the fusion capability already learned.

As shown in Fig.~\ref{fig:pipeline}, NatureFusion uses the pretrained generative model FLUX.2 [klein]~\cite{bfl2026flux2klein} as its backbone, with parameters $\theta_0$. We introduce two decoupled LoRA adapters~\cite{hu2022lora}, a fusion adapter with parameters $\phi_F$ and a task adapter with parameters $\phi_T$. Training proceeds in three stages. Stage I uses a generative foundation model as the teacher to produce fusion demonstrations $\tilde I_f$ offline from both inputs (Sec.~\ref{subsec:teacher_guidance}). Stage II trains the fusion adapter through flow matching, using these demonstrations as supervision and the image pair as the condition (Sec.~\ref{subsec:fusion_learning}). Stage III freezes the fusion adapter and trains only the task adapter to strengthen the complementary contribution of infrared under frozen detector supervision (Sec.~\ref{subsec:task_adaptation}). At inference, both adapters operate jointly, and only the image pair is required (Sec.~\ref{subsec:inference}). The person bounding boxes $\mathcal{B}$ are taken from dataset annotations and used only during training. They define infrared guidance regions for the teacher and supervise detection in Stage III.

\subsection{Teacher Guidance}
\label{subsec:teacher_guidance}

Because ideal fused images cannot be directly captured, we construct supervision using a generative foundation model as the teacher. The teacher $\mathcal{T}$ is OpenAI's GPT-Image-2, invoked in image editing mode. It operates offline before training, and its parameters are not optimized. To restrict infrared guidance to relevant regions, we retain infrared intensities only within the person bounding boxes to construct a guidance map,
\begin{equation}
G=M_{\mathcal{B}}\odot\bar I_{\mathrm{ir}},
\qquad
\tilde I_f=\mathcal{T}\big(I_{\mathrm{vi}},G;y_{\mathcal{T}}\big),
\label{eq:teacher}
\end{equation}
where $M_{\mathcal{B}}$ is a binary mask covering the union of the person boxes, $\bar I_{\mathrm{ir}}$ is the infrared intensity normalized to $[0,1]$, $\odot$ denotes pixelwise multiplication, and $y_{\mathcal{T}}$ is the teacher prompt. The prompt specifies the number of people and instructs the teacher to restore scene illumination and reveal people in dark regions using infrared cues while preserving the visible appearance and scene layout. Generation uses two successive calls. The first restores the full image and reveals the people. The second refines the background while keeping the people unchanged, producing the demonstration $\tilde I_f$. We run person detection on the output of the first call. If any people are missed, we supplement the prompt with their locations and repeat the call. Automatic quality checks then discard samples with structural inconsistencies relative to the inputs or abnormal exposure or color. This process yields training triplets $(I_{\mathrm{vi}},I_{\mathrm{ir}},\tilde I_f)$, with the person boxes $\mathcal{B}$ retained for Stage III.

\subsection{Fusion Learning}
\label{subsec:fusion_learning}

In Stage II, the fusion adapter learns image-pair-conditioned fusion from the demonstrations. The backbone's frozen VAE encoder $\mathcal{E}$ maps the demonstration and both inputs to a shared latent space,
\begin{equation}
z^\star=\mathcal{E}(\tilde I_f),\qquad
z_{\mathrm{vi}}=\mathcal{E}(I_{\mathrm{vi}}),\qquad
z_{\mathrm{ir}}=\mathcal{E}\big(\mathcal{P}_{\mathrm{ir}}(I_{\mathrm{ir}})\big),
\label{eq:latent_encoding}
\end{equation}
where $\mathcal{P}_{\mathrm{ir}}$ normalizes the full infrared image using its 1st and 99th intensity percentiles, clips the result to $[0,1]$, and replicates it across three channels. Unlike the teacher, the student receives the complete infrared image, and person boxes are excluded from its condition. All image pairs share a fixed text prompt $y$. It identifies the first reference as a low-light visible image and the second as a registered infrared image, requests a clear and natural visible image, and restricts the use of infrared to structural cues consistent with the visible image. The backbone's text encoder maps $y$ to $e_y$, which combines with the input latents to form the condition $c=(z_{\mathrm{vi}},z_{\mathrm{ir}},e_y)$. The two reference latents are flattened into token sequences and appended after the noisy target tokens in the order of visible followed by infrared. Source coordinates in the positional encoding distinguish the references. The backbone applies joint attention to all tokens, allowing target tokens to obtain appearance cues from visible tokens and structural cues from infrared tokens. This provides a mechanism for translating infrared cues into visible appearance. Only outputs corresponding to target tokens are supervised.

The fusion adapter modifies the backbone through low-rank weight increments. For each adapted linear layer,
\begin{equation}
W_F=W_0+\Delta W_F,\qquad
\Delta W_F=\frac{\alpha_F}{r_F}B_FA_F,
\label{eq:fusion_adapter}
\end{equation}
where $W_0\in\mathbb{R}^{d_{\mathrm{out}}\times d_{\mathrm{in}}}$ is a pretrained weight in $\theta_0$, $A_F\in\mathbb{R}^{r_F\times d_{\mathrm{in}}}$ and $B_F\in\mathbb{R}^{d_{\mathrm{out}}\times r_F}$ are low-rank matrices, and $r_F$ and $\alpha_F$ are the rank and scaling factor, respectively. The low-rank matrices across all adapted layers constitute $\phi_F$. We train the fusion adapter through flow matching~\cite{lipman2023flowmatching}. For each demonstration latent $z^\star$, we sample Gaussian noise $\epsilon\sim\mathcal{N}(0,\mathbf{I})$ and a noise level $\sigma$ uniformly from $(0,1]$ to construct the noisy target and target velocity,
\begin{equation}
z_\sigma=(1-\sigma)z^\star+\sigma\epsilon,
\qquad
u^\star=\epsilon-z^\star.
\label{eq:flow_interpolation}
\end{equation}
Let $v_F(z_\sigma,\sigma;c)$ denote the velocity predicted with only the fusion adapter enabled, abbreviated as $v_F(c)$ below. Stage II updates only $\phi_F$ by minimizing
\begin{equation}
\mathcal{L}_{\mathrm{FM}}=\mathbb{E}\Big[\big\|v_F(z_\sigma,\sigma;c)-u^\star\big\|^2\Big],
\label{eq:fusion_flow_loss}
\end{equation}
where $\|\cdot\|^2$ denotes squared error averaged over latent elements, and the expectation is over training samples, noise levels, and Gaussian noise. This procedure resembles offline distillation in form, but the student and teacher receive different inputs. The student is conditioned only on the image pair and fixed prompt, without access to the person boxes or counts used by the teacher. It must therefore infer from the complete infrared image where infrared cues are needed. Supervision consists solely of teacher output images and does not involve internal features or generation trajectories, imposing no requirement on the teacher architecture.

\subsection{Task Adaptation}
\label{subsec:task_adaptation}

Stage III introduces additional adjustments for downstream tasks while preserving the learned fusion capability. We freeze $\theta_0$ and $\phi_F$ and add a task adapter to the same linear layers,
\begin{equation}
W_{F+T}=W_0+\Delta W_F+\Delta W_T,\qquad
\Delta W_T=\frac{\alpha_T}{r_T}B_TA_T,
\label{eq:dual_adapter}
\end{equation}
where the symbols follow Eq.~\eqref{eq:fusion_adapter}. Only the task adapter parameters $\phi_T$ are updated. Task gradients do not propagate to $\phi_F$, so the fusion adapter learned in Stage II remains unchanged. We refer to the configuration with only the fusion adapter enabled as Fusion-only and the configuration with both adapters enabled as Fusion + Task. Their velocity predictions are denoted by $v_F$ and $v_{F+T}$, respectively. The subscript $F+T$ indicates that both adapters operate within the same backbone, rather than an addition of two predictions. We also replace the infrared input with a zero-valued image, whose latent is denoted by $z_\varnothing$, to construct the condition without infrared (w/o~IR), $c_\varnothing=(z_{\mathrm{vi}},z_\varnothing,e_y)$. It differs from $c$ only in the infrared input. All branches share the same noisy target $z_\sigma$ and noise level $\sigma$ constructed from the demonstration according to Eq.~\eqref{eq:flow_interpolation}. Differences between branches therefore arise solely from their conditions and active adapters. Each sample requires four forward passes, with the two Fusion-only passes computed without gradients.

\textbf{Detection loss.} We use pedestrian detection as the downstream task and backpropagate detection gradients through the generation process. The Fusion + Task velocity prediction provides an estimate of the clean latent at the current noise level, which is decoded by the frozen VAE decoder $\mathcal{D}$,
\begin{equation}
\hat z_f=z_\sigma-\sigma\,v_{F+T}(c),
\qquad
\hat I_f=\mathcal{D}(\hat z_f).
\label{eq:clean_estimate}
\end{equation}
The image $\hat I_f$ is a single-step estimate used during training, distinct from $I_f$ obtained through complete sampling. With a frozen detector $H_\psi$, the detection loss is
\begin{equation}
\mathcal{L}_{\mathrm{det}}=\ell_{\mathrm{det}}\big(H_\psi(\hat I_f),\mathcal{B}\big),
\label{eq:detection_loss}
\end{equation}
where $\ell_{\mathrm{det}}$ is the detector's native training loss, which supervises classification and localization using the person boxes. We use YOLO11x~\cite{jocher2024yolo11} pretrained on COCO~\cite{lin2014coco} as $H_\psi$. The detector and decoder remain frozen, but gradients propagate through both to $\phi_T$. Because single-step estimates are blurry at high noise levels, the detection loss is evaluated only for $\sigma\in[0.35,0.60]$, with the gate $g(\sigma)=\mathbf{1}[0.35\le\sigma\le0.60]$.

\textbf{Complementarity loss.} The detection loss constrains the correctness of detector outputs without distinguishing which modality supplies the discriminative cues. Optimizing this loss alone could encourage the task adapter to rely on textures unrelated to infrared to satisfy the detector. We therefore seek to guide its adjustments with infrared. For Fusion-only, the difference between predictions with and without infrared, $\Delta_{\mathrm{ir}}=v_F(c)-v_F(c_\varnothing)$, captures the information introduced by infrared beyond the visible image. We refer to this difference as the complementary contribution of infrared. Inspired by classifier-free guidance~\cite{ho2021cfg}, we extrapolate the Fusion-only prediction along $\Delta_{\mathrm{ir}}$ to construct the regression target for the task adapter,
\begin{equation}
v^\star=\operatorname{sg}\!\big[v_F(c_\varnothing)+w\,\Delta_{\mathrm{ir}}\big],
\quad
\mathcal{L}_{\mathrm{comp}}=\big\|v_{F+T}(c)-v^\star\big\|^2,
\label{eq:complementarity}
\end{equation}
where $\operatorname{sg}[\cdot]$ denotes stop-gradient and $w$ is the amplification factor. When $w=1$, $v^\star=v_F(c)$, and the target reduces to reproducing the Fusion-only prediction. When $w>1$, the complementary contribution is amplified. We use $w=2$.

\textbf{Anchoring loss.} The target in Eq.~\eqref{eq:complementarity} contains two components, the visible-only prediction $v_F(c_\varnothing)$ and the amplified complementary contribution $w\,\Delta_{\mathrm{ir}}$. The Fusion + Task prediction can be decomposed correspondingly into the prediction without infrared, $v_{F+T}(c_\varnothing)$, and the change induced by adding infrared, $v_{F+T}(c)-v_{F+T}(c_\varnothing)$. The complementarity loss constrains only their sum and cannot distinguish which component the task adapter modifies. Our objective is to modify only the complementary infrared contribution while preserving the visible-only component of Fusion-only, so that task gains build on infrared information. We therefore anchor the prediction without infrared to the corresponding Fusion-only prediction,
\begin{equation}
\mathcal{L}_{\mathrm{anc}}=\big\|v_{F+T}(c_\varnothing)-\operatorname{sg}\big[v_F(c_\varnothing)\big]\big\|^2.
\label{eq:anchoring}
\end{equation}
When both the complementarity and anchoring losses are zero, the two components match individually, yielding $v_{F+T}(c)-v_{F+T}(c_\varnothing)=w\,\Delta_{\mathrm{ir}}$. In this case, the task adapter exactly amplifies the complementary contribution of infrared by a factor of $w$. This term also penalizes adjustments induced by the detection loss that alter the visible-only component.

\textbf{Overall objective.} Stage III optimizes
\begin{equation}
\min_{\phi_T}\ \mathbb{E}\Big[\lambda_{\mathrm{det}}\,g(\sigma)\,\mathcal{L}_{\mathrm{det}}+\mathcal{L}_{\mathrm{comp}}+\lambda_{\mathrm{anc}}\mathcal{L}_{\mathrm{anc}}\Big],
\label{eq:task_total}
\end{equation}
where the expectation is over training samples, noise levels, and Gaussian noise. The detection loss is disabled for the first 50 updates. Its weight then increases linearly from 0 to $\lambda_{\mathrm{det}}=0.05$ over the next 100 updates, introducing detection supervision gradually. The complementarity and anchoring losses apply at all noise levels, with $\lambda_{\mathrm{anc}}=1$.

\subsection{Inference}
\label{subsec:inference}

Training uses FLUX.2 [klein] Base 9B, on which all preceding velocity predictions and training objectives are defined. At inference, we load both adapters onto FLUX.2 [klein] 9B from the same model family. This is the officially released model distilled for few-step sampling, hereafter referred to as the Distilled backbone. The two backbones share the same architecture, allowing the low-rank increments learned on the Base backbone to be loaded directly. Denoting the Distilled backbone parameters by $\theta_D$, the effective weight of each adapted linear layer at inference is
\begin{equation}
W=W_D+\Delta W_F+\Delta W_T,
\label{eq:inference_adapter}
\end{equation}
where $W_D$ is the corresponding weight in $\theta_D$. We denote the velocity prediction under these effective weights by $v_D$. Given $I_{\mathrm{vi}}$ and $I_{\mathrm{ir}}$, we construct the condition $c$ at the native input resolution according to Eq.~\eqref{eq:latent_encoding}. Starting from Gaussian noise $z_1$, we perform four Euler steps over the noise levels $1=\sigma_0>\sigma_1>\sigma_2>\sigma_3>\sigma_4=0$ specified by the official schedule,
\begin{equation}
z_{\sigma_{i+1}}=z_{\sigma_i}+(\sigma_{i+1}-\sigma_i)\,v_D(z_{\sigma_i},\sigma_i;c),\quad i=0,\dots,3,
\label{eq:sampling}
\end{equation}
and obtain the fused image as $I_f=\mathcal{D}(z_0)$. Inference requires only the registered image pair and fixed prompt, without the teacher, person boxes, or detector.

\section{Experiments}
\label{sec:experiments}

\subsection{Experimental Setup}
\label{subsec:experimental_setup}

\textbf{Datasets.} Following the procedure in Sec.~\ref{subsec:teacher_guidance}, we construct LLVIP-NCE from LLVIP~\cite{jia2021llvip}. It contains 779 sets of registered visible and infrared images, teacher fusion demonstrations, and person bounding box annotations. We divide them into 536 training, 132 validation, and 111 test samples with no scene overlap between splits. Both adapters are learned exclusively on the training split. The test split is used for component ablations, comparisons of teacher generation with and without infrared guidance, and teacher and student comparisons.

Cross-dataset evaluation uses SMOD~\cite{chen2025amfd}, MFNet~\cite{ha2017mfnet}, and M3FD~\cite{liu2022tardal}, none of which is used to train NatureFusion. Dataset suffixes denote the number of image pairs. Pedestrian detection and missed pedestrian recall are evaluated on SMOD-200, LLVIP-250, and M3FD-150. SMOD-200 is a fixed set of 200 nighttime SMOD pairs, with the original pedestrian and rider annotations merged into a single pedestrian category. LLVIP-250 is a fixed set of 250 LLVIP pairs. M3FD-150 consists of 150 pairs selected from M3FD. Fusion statistics and perceptual quality are evaluated on 20 pairs per dataset, randomly drawn from SMOD-200, MFNet-250, and M3FD-150, respectively, and shared by all methods. On all evaluation sets, every method is applied without further training, and no evaluator is fine-tuned on them.

Frozen segmentation is evaluated on MFNet-250, a fixed set of 250 MFNet pairs (76 daytime and 174 nighttime). For retrained segmentation, the segmenters are trained on 100 additional pairs outside MFNet-250. Removing the 59 pairs of MFNet-250 that share acquisition dates with these training pairs yields the test set MFNet-191. The 100 pairs are used only to train the segmenters, and all image generation methods, including NatureFusion, remain fixed.

\textbf{Compared methods.} We compare against fourteen methods. The seven multi-modal fusion methods are Text-IF~\cite{yi2024textif}, OmniFuse~\cite{zhang2025omnifuse}, HCLFuse~\cite{guo2025hclfuse}, DSPFusion~\cite{tang2026dspfusion}, DuS-DiFuse~\cite{cao2026dusdifuse}, DiTFuse~\cite{li2026ditfuse}, and ISFuse~\cite{wang2026isfuse}. The seven single-modality enhancement methods comprise six low-light enhancement methods, namely Retinexformer~\cite{cai2023retinexformer}, DiffLL~\cite{jiang2023diffll}, QuadPrior~\cite{wang2024quadprior}, DarkIR~\cite{feijoo2025darkir}, HVI-CIDNet~\cite{yan2025hvi}, and Diff-Retinex++~\cite{yi2025diffretinexpp}, together with the single-image fusion method MagicFuse~\cite{zhang2026magicfuse}. We use the public implementations and corresponding pretrained weights. Multi-modal fusion methods receive identical visible and infrared images, and single-modality enhancement methods receive the same visible input.

\textbf{Evaluation metrics.} We assess image quality through fusion statistics and perceptual quality. The fusion metrics are entropy (EN), spatial frequency (SF), standard deviation (SD), average gradient (AG), edge intensity (EI), and mutual information (MI), all computed on 8-bit grayscale outputs and reported with higher values indicating better performance. The first five characterize information content, contrast, or local variation in the output, whereas MI sums the mutual information between the output and each of the two source images, measuring its statistical dependence on their intensities. Perceptual quality is evaluated using six no-reference metrics, including DBCNN~\cite{zhang2020dbcnn}, MUSIQ~\cite{ke2021musiq}, MANIQA~\cite{yang2022maniqa}, ARNIQA~\cite{agnolucci2024arniqa}, CLIP-IQA~\cite{wang2023clipiqa}, and LIQE~\cite{zhang2023liqe}. Higher scores indicate better predicted quality for all six metrics. Ablations and teacher and student comparisons additionally report NIQE~\cite{mittal2013niqe}, for which lower values are better. All no-reference metrics use their corresponding default configurations in PyIQA 0.1.16~\cite{chen2022pyiqa}, taking RGB outputs as input and applying their respective internal preprocessing. DBCNN uses the KonIQ-10k weights~\cite{hosu2020koniq}. All image quality metrics are computed per image and then averaged.

Downstream evaluation primarily uses frozen models pretrained on visible images to assess whether fused images can be directly used by existing vision models. Detection is evaluated with seven COCO-pretrained models~\cite{lin2014coco}, including YOLO11x~\cite{jocher2024yolo11}, YOLO26x~\cite{ultralytics2026yolo26}, Faster R-CNN~\cite{ren2015fasterrcnn}, RetinaNet~\cite{lin2017focalloss}, Cascade R-CNN~\cite{cai2018cascadercnn}, RT-DETRv2~\cite{lv2024rtdetrv2}, and D-FINE-L~\cite{peng2025dfine}. RT-DETRv2 is abbreviated as RTDv2 in the tables. Only YOLO11x provides detection supervision in Stage III. The other six detectors do not participate in NatureFusion training. Following the COCO protocol, we compute pedestrian $\mathrm{AP}_{50:95}$ as the mean of ten AP values at IoU thresholds from 0.50 to 0.95 in increments of 0.05. We also report $\mathrm{AP}_{\tau}$ at individual IoU thresholds in the corresponding experiments. All detectors remain frozen, and each detector uses identical preprocessing and postprocessing across methods. Each detector computes AP over the full evaluation set, and Mean in the tables is the equal-weight average over all seven detectors. Unless otherwise specified, detection metrics in all figures and tables are averaged over the seven detectors.

To measure whether fused images recover targets missed in visible images, we define missed pedestrian recall $R_{\mathrm{miss}}$. On each original visible image, predicted boxes with confidence scores of at least 0.3 are matched one to one with ground-truth boxes in descending confidence order. A match requires an IoU of at least 0.5. Ground-truth pedestrians missed by all seven detectors form a fixed missed-target set. Using the same matching rule, we compute each detector's recall on this set for each method's output and then average over the seven detectors. $R_{\mathrm{miss}}$ is expressed as a percentage, with higher values indicating that a larger fraction of pedestrians missed in the visible images are subsequently detected.

Frozen segmentation uses seven Cityscapes-pretrained models~\cite{cordts2016cityscapes}, including PSPNet-R50~\cite{zhao2017pspnet}, DeepLabV3+ R50~\cite{chen2018deeplabv3plus}, OCRNet-HR48~\cite{yuan2020ocrnet}, SegFormer-B2~\cite{xie2021segformer}, Mask2Former-Swin-T~\cite{cheng2022mask2former}, SegMAN-B~\cite{fu2025segman}, and YOLO26x-sem~\cite{ultralytics2026yolo26}. Predictions and MFNet annotations are mapped to three common categories, Car, Person, and Other. We accumulate a pixel confusion matrix over the full evaluation set and compute Car IoU, Person IoU, and their mean $\mathrm{mIoU}_2$. Other is excluded from the class average, but pixels incorrectly assigned from Other to Car or Person contribute false positives to the respective category. Retrained segmentation additionally reports the mean IoU over all nine classes, including background, denoted $\mathrm{mIoU}_9$. All IoU scores are expressed as percentages, with higher values indicating better performance. DLV3+, Mask2F, and Y26-sem in the tables denote DeepLabV3+, Mask2Former, and YOLO26x-sem. In all figures and tables, Visible and Infrared are listed for reference and excluded from ranking.

\begin{figure*}[!t]
    \centering
    \includegraphics[width=\textwidth]{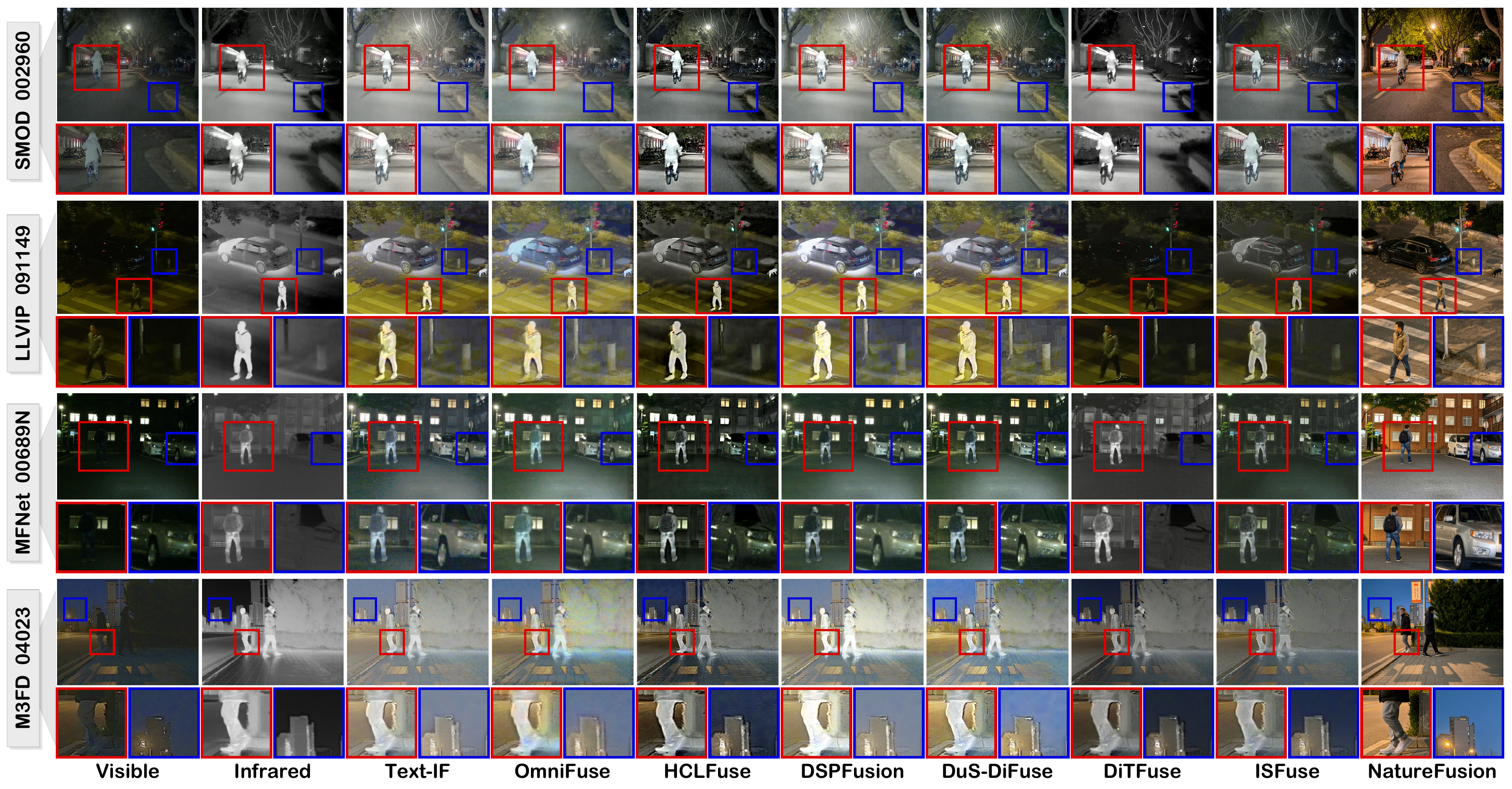}
    \caption{Qualitative comparison with multi-modal fusion methods.}
    \label{fig:qualitative_fusion}
\par\vspace{4pt}
\end{figure*}

\textbf{Implementation details.} Both adapters are trained on the frozen FLUX.2 [klein] Base 9B backbone with an input resolution of $640\times512$. In Stage II, the fusion adapter rank and scaling factor are set to $r_F=\alpha_F=16$, and training runs for 300 optimizer updates on a single RTX 3090 with a learning rate of $5\times10^{-5}$. Stage III freezes the fusion adapter and trains the task adapter with $r_T=\alpha_T=4$ for 300 optimizer updates at a learning rate of $2\times10^{-5}$. Both stages use 8-bit AdamW, BF16 mixed precision, a batch size of 1, and gradient accumulation over 4 steps. The learning rate is linearly warmed up over the first 50 updates, and the main model uses a fixed training seed of 42. Stage III uses $w=2$, $\lambda_{\mathrm{det}}=0.05$, and $\lambda_{\mathrm{anc}}=1$. The noise gating and gradual activation of the detection loss are described in Sec.~\ref{subsec:task_adaptation}.

\subsection{Comparison With Multi-Modal Image Fusion Methods}
\label{subsec:fusion_comparison}

This section evaluates NatureFusion in terms of fusion statistics, natural appearance, and target presentation against the seven multi-modal fusion methods listed in Sec.~\ref{subsec:experimental_setup}.

\textbf{Qualitative comparison.} Fig.~\ref{fig:qualitative_fusion} presents four nighttime scenes from SMOD, LLVIP, MFNet, and M3FD, with target and background regions enlarged. Text-IF, OmniFuse, HCLFuse, DSPFusion, DuS-DiFuse, and ISFuse predominantly render pedestrians with the bright appearance of the infrared input. For example, large areas of the rider's upper body in SMOD and the pedestrian's clothing and legs in M3FD are washed out, reducing visual distinctions between clothing and body parts. Similar effects occur in the background. Light patches appear on the originally dark vehicle body in LLVIP, while dark vegetation in M3FD appears grayish white or yellowish. DiTFuse better preserves the visible appearance in the LLVIP scene, but the pedestrian in the dark remains indistinct. Its outputs in the SMOD and MFNet scenes are nearly grayscale.

NatureFusion renders these targets with visible appearance consistent with the scene. The rider's jacket and bicycle in SMOD, the pedestrian's jacket and trousers in LLVIP, the backpack in MFNet, and the pedestrian's dark clothing and shoes in M3FD exhibit distinct visual details, with target locations and poses corresponding to the inputs. The curbs, lamp post base and bollard, vehicle front, and distant building in the blue boxes also exhibit clearer structure and more coherent colors. These results illustrate the integration of infrared structural cues with visible appearance.

\textbf{Fusion metrics.} Table~\ref{tab:fusion_comparison} reports the mean values of six fusion metrics on the evaluation subsets. NatureFusion ranks first in EN, SF, SD, AG, and EI on all three datasets, yielding 15 best results and indicating high information content, contrast, and local gradient responses. Relative to the second-best method for each metric, its margins in SF, AG, and EI are largest on MFNet and smaller on M3FD. The MI ranking differs. DSPFusion ranks first on all three datasets, whereas NatureFusion does not rank highly. This difference also reflects a characteristic of our approach. MI measures intensity-based statistical dependence between the output and source images, while translating infrared cues into natural appearance changes their intensity relationships in the output. The results indicate that source fidelity and natural appearance quality are not equivalent.

\begin{table*}[!t]
\centering
\makebox[\linewidth][c]{\begin{minipage}{\textwidth}
\centering
\caption{Fusion metric comparison with multi-modal fusion methods. The best and second-best values in each column are marked in bold red and underlined blue, respectively.}
\label{tab:fusion_comparison}
\fontsize{7.5}{9}\selectfont
\setlength{\tabcolsep}{1pt}
\renewcommand{\arraystretch}{1.16}
\begin{tabular*}{\textwidth}{@{\extracolsep{\fill}}lcccccccccccccccccc@{}}
\toprule
\multirow{2}{*}[-1.5pt]{Method} & \multicolumn{6}{c}{SMOD} & \multicolumn{6}{c}{MFNet} & \multicolumn{6}{c}{M3FD} \\
\cmidrule(lr){2-7}\cmidrule(lr){8-13}\cmidrule(lr){14-19}
 & EN$\uparrow$ & SF$\uparrow$ & SD$\uparrow$ & AG$\uparrow$ & EI$\uparrow$ & MI$\uparrow$ & EN$\uparrow$ & SF$\uparrow$ & SD$\uparrow$ & AG$\uparrow$ & EI$\uparrow$ & MI$\uparrow$ & EN$\uparrow$ & SF$\uparrow$ & SD$\uparrow$ & AG$\uparrow$ & EI$\uparrow$ & MI$\uparrow$ \\
\midrule
Text-IF~\cite{yi2024textif} & 7.202 & 10.730 & 41.075 & 3.479 & 36.046 & 3.288 & 6.809 & 8.781 & 37.402 & 3.095 & 33.312 & \textcolor[RGB]{0,0,220}{\underline{3.212}} & 6.390 & 9.041 & 24.737 & 2.787 & 29.407 & 2.535 \\
OmniFuse~\cite{zhang2025omnifuse} & 7.113 & 7.265 & 37.892 & 2.291 & 25.092 & 3.225 & 6.873 & 8.705 & 40.825 & 2.751 & 30.370 & 2.449 & 6.517 & 8.135 & 30.190 & 2.400 & 26.134 & \textcolor[RGB]{0,0,220}{\underline{3.178}} \\
HCLFuse~\cite{guo2025hclfuse} & \textcolor[RGB]{0,0,220}{\underline{7.299}} & \textcolor[RGB]{0,0,220}{\underline{19.526}} & 47.552 & \textcolor[RGB]{0,0,220}{\underline{6.265}} & \textcolor[RGB]{0,0,220}{\underline{63.880}} & 2.444 & 6.476 & \textcolor[RGB]{0,0,220}{\underline{14.154}} & 40.466 & \textcolor[RGB]{0,0,220}{\underline{4.495}} & \textcolor[RGB]{0,0,220}{\underline{47.276}} & 1.875 & 7.085 & 17.331 & 38.223 & 5.903 & \textcolor[RGB]{0,0,220}{\underline{61.154}} & 2.035 \\
DSPFusion~\cite{tang2026dspfusion} & 6.991 & 11.659 & 36.877 & 3.515 & 36.236 & \textcolor[RGB]{220,0,0}{\textbf{3.994}} & 6.961 & 10.822 & \textcolor[RGB]{0,0,220}{\underline{46.086}} & 3.515 & 37.679 & \textcolor[RGB]{220,0,0}{\textbf{3.855}} & 6.589 & 12.585 & 32.006 & 3.872 & 40.184 & \textcolor[RGB]{220,0,0}{\textbf{3.476}} \\
DuS-DiFuse~\cite{cao2026dusdifuse} & 7.263 & 16.246 & 42.466 & 4.758 & 48.570 & 2.955 & \textcolor[RGB]{0,0,220}{\underline{7.075}} & 13.841 & 45.815 & 4.459 & 46.834 & 2.491 & \textcolor[RGB]{0,0,220}{\underline{7.154}} & \textcolor[RGB]{0,0,220}{\underline{19.308}} & \textcolor[RGB]{0,0,220}{\underline{38.995}} & \textcolor[RGB]{0,0,220}{\underline{5.908}} & 60.775 & 1.943 \\
DiTFuse~\cite{li2026ditfuse} & 7.151 & 10.041 & \textcolor[RGB]{0,0,220}{\underline{56.110}} & 3.061 & 32.493 & \textcolor[RGB]{0,0,220}{\underline{3.840}} & 6.196 & 7.000 & 31.091 & 2.245 & 24.514 & 2.345 & 6.859 & 10.638 & 33.396 & 3.165 & 33.909 & 2.334 \\
ISFuse~\cite{wang2026isfuse} & 7.161 & 13.912 & 43.288 & 4.205 & 42.063 & 2.883 & 6.238 & 9.699 & 30.507 & 3.245 & 33.807 & 2.237 & 6.871 & 12.239 & 32.663 & 3.876 & 39.482 & 2.546 \\
\midrule
\textbf{NatureFusion} & \textcolor[RGB]{220,0,0}{\textbf{7.597}} & \textcolor[RGB]{220,0,0}{\textbf{26.675}} & \textcolor[RGB]{220,0,0}{\textbf{56.587}} & \textcolor[RGB]{220,0,0}{\textbf{9.135}} & \textcolor[RGB]{220,0,0}{\textbf{88.398}} & 1.850 & \textcolor[RGB]{220,0,0}{\textbf{7.537}} & \textcolor[RGB]{220,0,0}{\textbf{24.584}} & \textcolor[RGB]{220,0,0}{\textbf{57.578}} & \textcolor[RGB]{220,0,0}{\textbf{8.212}} & \textcolor[RGB]{220,0,0}{\textbf{80.484}} & 1.649 & \textcolor[RGB]{220,0,0}{\textbf{7.447}} & \textcolor[RGB]{220,0,0}{\textbf{20.655}} & \textcolor[RGB]{220,0,0}{\textbf{53.562}} & \textcolor[RGB]{220,0,0}{\textbf{6.306}} & \textcolor[RGB]{220,0,0}{\textbf{62.102}} & 1.617 \\
\bottomrule
\end{tabular*}
\end{minipage}}
\end{table*}

\textbf{Perceptual quality.} Table~\ref{tab:perceptual_comparison} reports six no-reference perceptual metrics on the same samples. NatureFusion obtains the best score on all metrics across the three datasets, and DuS-DiFuse ranks second in most cases. Under the source-fidelity paradigm, the dark visible background and bright infrared targets are carried together into the output, producing the washed-out targets and grayish backgrounds in the compared results in Fig.~\ref{fig:qualitative_fusion}. NatureFusion instead converts infrared cues into visible appearance, achieving its largest advantage on the darker MFNet scenes, where its MUSIQ score is nearly twice the second-best. Although the compared methods also take the infrared image as input, this comparison shows that access to infrared alone does not guarantee perceptual quality. What matters is how infrared information is presented in the output.

\begin{table*}[!t]
\centering
\makebox[\linewidth][c]{\begin{minipage}{\textwidth}
\centering
\caption{No-reference perceptual comparison with multi-modal fusion methods.}
\label{tab:perceptual_comparison}
\fontsize{8}{9.6}\selectfont
\setlength{\tabcolsep}{0.35pt}
\renewcommand{\arraystretch}{1.2}
\begin{tabular*}{\textwidth}{@{\extracolsep{\fill}}lcccccccccccccccccc@{}}
\toprule
\multirow{2}{*}[-1.5pt]{Method} & \multicolumn{6}{c}{SMOD Dataset} & \multicolumn{6}{c}{MFNet Dataset} & \multicolumn{6}{c}{M3FD Dataset} \\
\cmidrule(lr){2-7}\cmidrule(lr){8-13}\cmidrule(lr){14-19}
 & {\fontsize{6.5}{7.8}\selectfont DBCNN} & {\fontsize{6.5}{7.8}\selectfont MUSIQ} & {\fontsize{6.5}{7.8}\selectfont MANIQA} & {\fontsize{6.5}{7.8}\selectfont ARNIQA} & {\fontsize{6.5}{7.8}\selectfont CLIP-IQA} & {\fontsize{6.5}{7.8}\selectfont LIQE} & {\fontsize{6.5}{7.8}\selectfont DBCNN} & {\fontsize{6.5}{7.8}\selectfont MUSIQ} & {\fontsize{6.5}{7.8}\selectfont MANIQA} & {\fontsize{6.5}{7.8}\selectfont ARNIQA} & {\fontsize{6.5}{7.8}\selectfont CLIP-IQA} & {\fontsize{6.5}{7.8}\selectfont LIQE} & {\fontsize{6.5}{7.8}\selectfont DBCNN} & {\fontsize{6.5}{7.8}\selectfont MUSIQ} & {\fontsize{6.5}{7.8}\selectfont MANIQA} & {\fontsize{6.5}{7.8}\selectfont ARNIQA} & {\fontsize{6.5}{7.8}\selectfont CLIP-IQA} & {\fontsize{6.5}{7.8}\selectfont LIQE} \\
\midrule
Text-IF~\cite{yi2024textif} & 0.443 & 55.879 & 0.245 & 0.584 & 0.229 & 1.210 & 0.214 & 30.951 & 0.089 & 0.396 & 0.109 & 1.023 & 0.402 & 49.077 & 0.257 & 0.561 & 0.211 & 1.146 \\
OmniFuse~\cite{zhang2025omnifuse} & 0.328 & 40.169 & 0.159 & 0.514 & 0.233 & 1.007 & \textcolor[RGB]{0,0,220}{\underline{0.300}} & 29.491 & 0.108 & 0.454 & \textcolor[RGB]{0,0,220}{\underline{0.190}} & 1.036 & 0.375 & 46.566 & 0.255 & 0.506 & 0.323 & 1.032 \\
HCLFuse~\cite{guo2025hclfuse} & 0.403 & 54.967 & 0.233 & 0.583 & 0.199 & 1.262 & 0.222 & 34.024 & 0.100 & 0.462 & 0.136 & 1.019 & 0.348 & 47.018 & 0.224 & 0.489 & 0.244 & 1.077 \\
DSPFusion~\cite{tang2026dspfusion} & 0.441 & 55.543 & 0.256 & 0.570 & 0.247 & 1.362 & 0.207 & 32.265 & 0.094 & 0.446 & 0.147 & 1.043 & 0.442 & 52.768 & 0.250 & 0.539 & 0.232 & 1.172 \\
DuS-DiFuse~\cite{cao2026dusdifuse} & \textcolor[RGB]{0,0,220}{\underline{0.544}} & \textcolor[RGB]{0,0,220}{\underline{64.242}} & \textcolor[RGB]{0,0,220}{\underline{0.279}} & \textcolor[RGB]{0,0,220}{\underline{0.630}} & \textcolor[RGB]{0,0,220}{\underline{0.296}} & \textcolor[RGB]{0,0,220}{\underline{1.710}} & 0.271 & \textcolor[RGB]{0,0,220}{\underline{36.971}} & 0.115 & \textcolor[RGB]{0,0,220}{\underline{0.527}} & 0.170 & \textcolor[RGB]{0,0,220}{\underline{1.170}} & \textcolor[RGB]{0,0,220}{\underline{0.565}} & \textcolor[RGB]{0,0,220}{\underline{58.583}} & \textcolor[RGB]{0,0,220}{\underline{0.288}} & \textcolor[RGB]{0,0,220}{\underline{0.590}} & \textcolor[RGB]{0,0,220}{\underline{0.391}} & \textcolor[RGB]{0,0,220}{\underline{1.523}} \\
DiTFuse~\cite{li2026ditfuse} & 0.350 & 50.633 & 0.269 & 0.504 & 0.245 & 1.127 & 0.275 & 33.640 & \textcolor[RGB]{0,0,220}{\underline{0.161}} & 0.444 & 0.129 & 1.022 & 0.398 & 51.058 & 0.271 & 0.570 & 0.263 & 1.206 \\
ISFuse~\cite{wang2026isfuse} & 0.499 & 55.887 & 0.275 & 0.589 & 0.183 & 1.244 & 0.250 & 34.262 & 0.115 & 0.436 & 0.143 & 1.021 & 0.457 & 52.473 & 0.250 & 0.525 & 0.230 & 1.148 \\
\midrule
\textbf{NatureFusion} & \textcolor[RGB]{220,0,0}{\textbf{0.645}} & \textcolor[RGB]{220,0,0}{\textbf{71.960}} & \textcolor[RGB]{220,0,0}{\textbf{0.461}} & \textcolor[RGB]{220,0,0}{\textbf{0.713}} & \textcolor[RGB]{220,0,0}{\textbf{0.601}} & \textcolor[RGB]{220,0,0}{\textbf{4.064}} & \textcolor[RGB]{220,0,0}{\textbf{0.688}} & \textcolor[RGB]{220,0,0}{\textbf{69.421}} & \textcolor[RGB]{220,0,0}{\textbf{0.442}} & \textcolor[RGB]{220,0,0}{\textbf{0.719}} & \textcolor[RGB]{220,0,0}{\textbf{0.587}} & \textcolor[RGB]{220,0,0}{\textbf{4.669}} & \textcolor[RGB]{220,0,0}{\textbf{0.690}} & \textcolor[RGB]{220,0,0}{\textbf{73.731}} & \textcolor[RGB]{220,0,0}{\textbf{0.442}} & \textcolor[RGB]{220,0,0}{\textbf{0.717}} & \textcolor[RGB]{220,0,0}{\textbf{0.590}} & \textcolor[RGB]{220,0,0}{\textbf{4.283}} \\
\bottomrule
\end{tabular*}
\end{minipage}}
\par\vspace{4pt}
\end{table*}

\subsection{Comparison With Single-Modality Image Enhancement Methods}
\label{subsec:lle_comparison}

This section examines whether single-modality enhancement can improve both the visual quality of nighttime scenes and the discernibility of targets in dark regions, and what additional information infrared can provide. We compare against the six low-light enhancement methods and the single-image fusion method MagicFuse listed in Sec.~\ref{subsec:experimental_setup}. Evaluation addresses two aspects of natural complementary enhancement, using no-reference perceptual metrics for visual quality and $R_{\mathrm{miss}}$ for the recovery of pedestrians missed in visible images.

\begin{figure*}[!t]
    \centering
    \includegraphics[width=\textwidth]{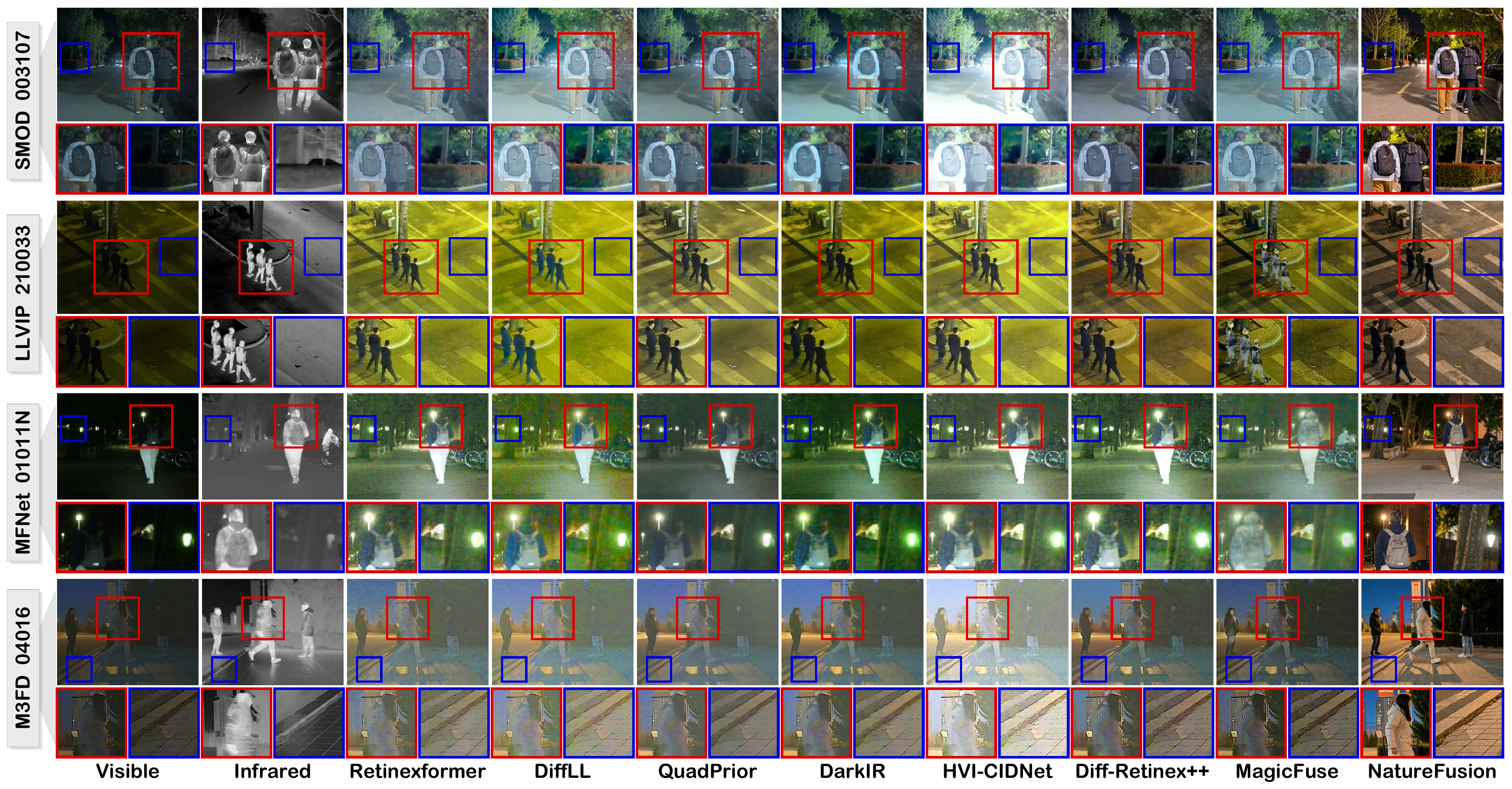}
    \caption{Qualitative comparison with single-modality enhancement methods.}
    \label{fig:qualitative_lle}
\par\vspace{3pt}
\end{figure*}

\textbf{Qualitative comparison.} Fig.~\ref{fig:qualitative_lle} shows nighttime pedestrian scenes from four datasets, with person and background regions enlarged. In these scenes, brightness improvement often accompanies residual color casts, noise, or local overexposure. In the LLVIP scene, all six low-light enhancement methods retain the yellow cast caused by streetlights. The visible image in the MFNet scene is dark overall. The enhanced backgrounds exhibit pronounced chromatic noise, while large areas of the pedestrian's light trousers appear uniformly white, obscuring leg contours and folds. In SMOD, HVI-CIDNet washes out the left pedestrian's jacket and the road surface. MagicFuse renders pedestrians with a grayish white appearance in LLVIP and MFNet. In M3FD, the right pedestrian is almost concealed in the shadows of the vegetation in the visible image, and only the light trousers can be faintly distinguished in the outputs of the seven methods.

NatureFusion reduces the yellow cast in the LLVIP scene, producing more consistent tonal relationships between the three pedestrians' dark suits and the crosswalk. In SMOD, the pedestrian's backpack, jacket, and trousers are clearly differentiated. In MFNet, the backpack, dark jacket, and trouser folds are readily discernible. For the right pedestrian in M3FD, NatureFusion uses the infrared contour to render a more complete person in a dark jacket and light trousers.

\textbf{Perceptual quality.} Table~\ref{tab:lle_perceptual_comparison} shows that NatureFusion obtains the best score on all six no-reference metrics across the three datasets. The compared single-modality enhancement methods can only use the visible signal, and brightening dark inputs amplifies noise, as seen in the chromatic noise and overexposed trousers in the MFNet scene of Fig.~\ref{fig:qualitative_lle}. In contrast, NatureFusion consistently draws complementary information from infrared images to supply the target structure missing in low-light visible images, maintaining stable performance across datasets. The difference between the compared methods and NatureFusion is particularly pronounced in LIQE, which scores images by measuring their correspondence with text descriptions of different quality levels. On its scale from 1 to 5, the compared methods score between 1.0 and 1.4, close to the lowest level, whereas NatureFusion reaches 4.1 to 4.7. Since such learned metrics derive their quality judgments from natural photographs, this gap also indicates that NatureFusion produces outputs closer to well-exposed natural images.

\textbf{Recall of pedestrians missed in visible images.} Table~\ref{tab:lle_detection_comparison} reports $\mathrm{AP}_{50:95}$ and $R_{\mathrm{miss}}$ averaged over seven frozen detectors on SMOD-200, LLVIP-250, and M3FD-150. Although the seven single-modality enhancement methods use different enhancement strategies, their detection performance remains similar. Across the datasets, the maximum spread in $\mathrm{AP}_{50:95}$ is only about 7 percentage points, and their $R_{\mathrm{miss}}$ values remain below 5\%, falling below 1\% on M3FD. In other words, although single-modality enhancement changes the appearance of visible images, it rarely makes missed pedestrians detectable, because targets hidden in darkness leave too little visible signal for any enhancement prior to recover. NatureFusion recovers 32.5\% to 47.1\% of these pedestrians and achieves substantially higher $\mathrm{AP}_{50:95}$ than the compared methods on all three datasets, with the largest margins on LLVIP and M3FD. This gain comes from exploiting complementary information rather than increasing enhancement strength. Infrared supplies the locations and contours of these targets, and NatureFusion renders them with a natural appearance that detectors pretrained on visible images can recognize.

\begin{table*}[!t]
\centering
\makebox[\linewidth][c]{\begin{minipage}{\textwidth}
\centering

\caption{No-reference perceptual comparison with single-modality enhancement methods.}

\label{tab:lle_perceptual_comparison}
\fontsize{7.7}{9.6}\selectfont
\setlength{\tabcolsep}{0.4pt}
\renewcommand{\arraystretch}{1.2}
\begin{tabular*}{\textwidth}{@{\extracolsep{\fill}}lcccccccccccccccccc@{}}
\toprule
\multirow{2}{*}[-1.5pt]{Method} & \multicolumn{6}{c}{SMOD Dataset} & \multicolumn{6}{c}{MFNet Dataset} & \multicolumn{6}{c}{M3FD Dataset} \\
\cmidrule(lr){2-7}\cmidrule(lr){8-13}\cmidrule(lr){14-19}
 & {\fontsize{6.2}{7.8}\selectfont DBCNN} & {\fontsize{6.2}{7.8}\selectfont MUSIQ} & {\fontsize{6.2}{7.8}\selectfont MANIQA} & {\fontsize{6.2}{7.8}\selectfont ARNIQA} & {\fontsize{6.2}{7.8}\selectfont CLIP-IQA} & {\fontsize{6.2}{7.8}\selectfont LIQE} & {\fontsize{6.2}{7.8}\selectfont DBCNN} & {\fontsize{6.2}{7.8}\selectfont MUSIQ} & {\fontsize{6.2}{7.8}\selectfont MANIQA} & {\fontsize{6.2}{7.8}\selectfont ARNIQA} & {\fontsize{6.2}{7.8}\selectfont CLIP-IQA} & {\fontsize{6.2}{7.8}\selectfont LIQE} & {\fontsize{6.2}{7.8}\selectfont DBCNN} & {\fontsize{6.2}{7.8}\selectfont MUSIQ} & {\fontsize{6.2}{7.8}\selectfont MANIQA} & {\fontsize{6.2}{7.8}\selectfont ARNIQA} & {\fontsize{6.2}{7.8}\selectfont CLIP-IQA} & {\fontsize{6.2}{7.8}\selectfont LIQE} \\
\midrule
Retinexformer~\cite{cai2023retinexformer} & 0.399 & 48.570 & 0.207 & 0.592 & 0.107 & 1.031 & 0.242 & 29.354 & 0.068 & 0.459 & 0.081 & 1.018 & 0.451 & 47.704 & \textcolor[RGB]{0,0,220}{\underline{0.294}} & 0.571 & 0.249 & 1.383 \\
DiffLL~\cite{jiang2023diffll} & \textcolor[RGB]{0,0,220}{\underline{0.505}} & 46.944 & 0.166 & 0.570 & 0.094 & 1.037 & \textcolor[RGB]{0,0,220}{\underline{0.310}} & 27.548 & 0.060 & 0.414 & 0.109 & 1.028 & \textcolor[RGB]{0,0,220}{\underline{0.560}} & 54.594 & 0.243 & 0.531 & 0.270 & 1.208 \\
QuadPrior~\cite{wang2024quadprior} & 0.408 & 52.390 & 0.226 & 0.576 & 0.179 & 1.109 & 0.191 & 28.377 & 0.079 & 0.450 & 0.128 & 1.058 & 0.398 & 50.389 & 0.278 & 0.553 & \textcolor[RGB]{0,0,220}{\underline{0.391}} & 1.216 \\
DarkIR~\cite{feijoo2025darkir} & 0.365 & 50.923 & 0.221 & 0.546 & \textcolor[RGB]{0,0,220}{\underline{0.269}} & 1.058 & 0.210 & \textcolor[RGB]{0,0,220}{\underline{34.874}} & \textcolor[RGB]{0,0,220}{\underline{0.099}} & 0.452 & \textcolor[RGB]{0,0,220}{\underline{0.176}} & 1.029 & 0.453 & 55.075 & 0.290 & 0.547 & 0.365 & 1.264 \\
HVI-CIDNet~\cite{yan2025hvi} & 0.360 & 46.588 & 0.183 & 0.553 & 0.125 & 1.025 & 0.230 & 26.085 & 0.057 & 0.438 & 0.098 & 1.006 & 0.448 & 52.958 & 0.252 & 0.531 & 0.240 & \textcolor[RGB]{0,0,220}{\underline{1.403}} \\
Diff-Retinex++~\cite{yi2025diffretinexpp} & 0.451 & \textcolor[RGB]{0,0,220}{\underline{54.644}} & \textcolor[RGB]{0,0,220}{\underline{0.257}} & \textcolor[RGB]{0,0,220}{\underline{0.621}} & 0.213 & \textcolor[RGB]{0,0,220}{\underline{1.288}} & 0.264 & 29.268 & 0.082 & 0.459 & 0.144 & \textcolor[RGB]{0,0,220}{\underline{1.186}} & 0.490 & 55.776 & 0.271 & \textcolor[RGB]{0,0,220}{\underline{0.603}} & 0.187 & 1.304 \\
MagicFuse~\cite{zhang2026magicfuse} & 0.322 & 44.471 & 0.143 & 0.582 & 0.114 & 1.018 & 0.206 & 27.689 & 0.085 & \textcolor[RGB]{0,0,220}{\underline{0.483}} & 0.127 & 1.049 & 0.452 & \textcolor[RGB]{0,0,220}{\underline{56.301}} & 0.252 & 0.542 & 0.265 & 1.187 \\
\midrule
\textbf{NatureFusion} & \textcolor[RGB]{220,0,0}{\textbf{0.645}} & \textcolor[RGB]{220,0,0}{\textbf{71.960}} & \textcolor[RGB]{220,0,0}{\textbf{0.461}} & \textcolor[RGB]{220,0,0}{\textbf{0.713}} & \textcolor[RGB]{220,0,0}{\textbf{0.601}} & \textcolor[RGB]{220,0,0}{\textbf{4.064}} & \textcolor[RGB]{220,0,0}{\textbf{0.688}} & \textcolor[RGB]{220,0,0}{\textbf{69.421}} & \textcolor[RGB]{220,0,0}{\textbf{0.442}} & \textcolor[RGB]{220,0,0}{\textbf{0.719}} & \textcolor[RGB]{220,0,0}{\textbf{0.587}} & \textcolor[RGB]{220,0,0}{\textbf{4.669}} & \textcolor[RGB]{220,0,0}{\textbf{0.690}} & \textcolor[RGB]{220,0,0}{\textbf{73.731}} & \textcolor[RGB]{220,0,0}{\textbf{0.442}} & \textcolor[RGB]{220,0,0}{\textbf{0.717}} & \textcolor[RGB]{220,0,0}{\textbf{0.590}} & \textcolor[RGB]{220,0,0}{\textbf{4.283}} \\
\bottomrule
\end{tabular*}
\end{minipage}}
\end{table*}

\subsection{Downstream Vision Tasks}
\label{subsec:downstream}

This section evaluates whether fused images support object detection and semantic segmentation. We first fix the weights of evaluators pretrained on visible images and apply the same evaluation procedure to the seven multi-modal fusion methods and seven single-modality enhancement methods, using the original visible and infrared images as references. We then retrain segmenters separately on each method's outputs to assess performance after the segmenters adapt to the corresponding input distribution.

\textbf{Frozen detection.} Pedestrian detection on SMOD-200 and LLVIP-250 is evaluated with the seven COCO-pretrained detectors listed in Sec.~\ref{subsec:experimental_setup}. YOLO11x also supplies task supervision in Stage III, while the remaining six detectors assess whether detection gains transfer to models that do not participate in task adaptation.

\begin{figure*}[!t]
\centering
\includegraphics[width=\textwidth]{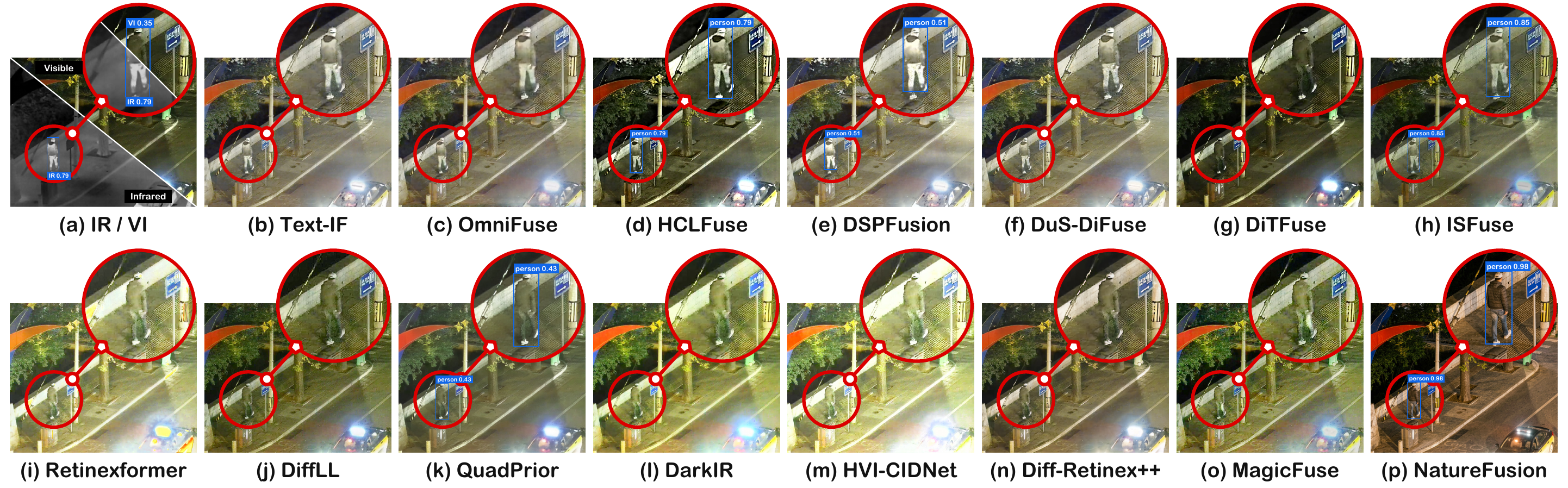}
\caption{Person detections from the same frozen RetinaNet on an LLVIP scene. Blue boxes and numbers show predictions with confidence of at least 0.25. (a) Visible (upper right) and infrared (lower left) images are detected separately and then combined. Red circles enlarge the person region.}
\label{fig:qualitative_detection}
\par\vspace{2pt}
\end{figure*}

\begin{figure}[!t]
\centering
\includegraphics[width=\columnwidth]{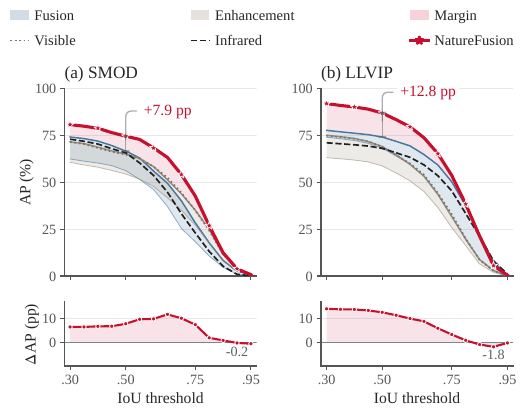}
\caption{Mean person AP versus IoU threshold. Fusion and Enhancement bands span the minimum to maximum AP of the multi-modal fusion and single-modality enhancement methods, respectively, and Margin shades where NatureFusion exceeds the best of them. Lower panels plot $\Delta$AP, NatureFusion minus the best compared method.}

\label{fig:det_iou_profile}
\end{figure}

\begin{table}[!t]
\centering
\begin{minipage}{\columnwidth}
\centering
\caption{Detection comparison with single-modality enhancement methods.}
\label{tab:lle_detection_comparison}
\fontsize{8}{9.6}\selectfont
\setlength{\tabcolsep}{0.4pt}
\renewcommand{\arraystretch}{1.16}
\begin{tabular*}{\columnwidth}{@{\extracolsep{\fill}}lcccccc@{}}
\toprule
\multirow{2}{*}[-1.5pt]{Method} & \multicolumn{2}{c}{SMOD} & \multicolumn{2}{c}{LLVIP} & \multicolumn{2}{c}{M3FD} \\
\cmidrule(lr){2-3}\cmidrule(lr){4-5}\cmidrule(lr){6-7}
 & {\fontsize{6.5}{7.8}\selectfont $\mathrm{AP}_{50:95}\uparrow$} & {\fontsize{6.5}{7.8}\selectfont $R_{\mathrm{miss}}\uparrow$} & {\fontsize{6.5}{7.8}\selectfont $\mathrm{AP}_{50:95}\uparrow$} & {\fontsize{6.5}{7.8}\selectfont $R_{\mathrm{miss}}\uparrow$} & {\fontsize{6.5}{7.8}\selectfont $\mathrm{AP}_{50:95}\uparrow$} & {\fontsize{6.5}{7.8}\selectfont $R_{\mathrm{miss}}\uparrow$} \\
\midrule
Retinexformer~\cite{cai2023retinexformer} & 34.03 & 2.4 & 31.73 & 1.9 & 30.51 & 0.0 \\
DiffLL~\cite{jiang2023diffll} & 29.82 & 0.5 & 29.72 & 2.6 & 33.64 & 0.3 \\
QuadPrior~\cite{wang2024quadprior} & 34.87 & 1.3 & \textcolor[RGB]{0,0,220}{\underline{35.31}} & 2.7 & \textcolor[RGB]{0,0,220}{\underline{35.15}} & 0.0 \\
DarkIR~\cite{feijoo2025darkir} & \textcolor[RGB]{0,0,220}{\underline{35.75}} & \textcolor[RGB]{0,0,220}{\underline{3.4}} & 33.92 & 3.0 & 33.64 & 0.1 \\
HVI-CIDNet~\cite{yan2025hvi} & 34.45 & 2.4 & 30.70 & 1.9 & 33.93 & 0.2 \\
Diff-Retinex++~\cite{yi2025diffretinexpp} & 35.02 & 2.1 & 31.92 & 2.9 & 33.78 & 0.3 \\
MagicFuse~\cite{zhang2026magicfuse} & 28.71 & 1.9 & 31.24 & \textcolor[RGB]{0,0,220}{\underline{4.9}} & 35.09 & \textcolor[RGB]{0,0,220}{\underline{0.9}} \\
\midrule
\textbf{NatureFusion} & \textcolor[RGB]{220,0,0}{\textbf{41.95}} & \textcolor[RGB]{220,0,0}{\textbf{32.5}} & \textcolor[RGB]{220,0,0}{\textbf{50.79}} & \textcolor[RGB]{220,0,0}{\textbf{47.1}} & \textcolor[RGB]{220,0,0}{\textbf{52.96}} & \textcolor[RGB]{220,0,0}{\textbf{45.0}} \\
\bottomrule
\end{tabular*}
\end{minipage}
\end{table}

\begin{table*}[!t]
\centering
\makebox[\linewidth][c]{\begin{minipage}{\textwidth}
\centering
\caption{Frozen detection comparison (\%). $\dagger$ marks YOLO11x, which provides Stage III supervision.}
\label{tab:detection_comparison}
\begingroup
\fontsize{7.5}{9}\selectfont
\setlength{\tabcolsep}{0.8pt}
\renewcommand{\arraystretch}{1.16}
\begin{tabular*}{\textwidth}{@{\extracolsep{\fill}}lcccccccccccccccccc@{}}
\toprule
\multirow{3.8}{*}[-3.5pt]{Method} & \multicolumn{9}{c}{SMOD} & \multicolumn{9}{c}{LLVIP} \\
\cmidrule(lr){2-10}\cmidrule(lr){11-19}
 & \multicolumn{8}{c}{$\mathrm{AP}_{50:95}\uparrow$} & $\mathrm{AP}_{50}\uparrow$ & \multicolumn{8}{c}{$\mathrm{AP}_{50:95}\uparrow$} & $\mathrm{AP}_{50}\uparrow$ \\
\cmidrule(lr){2-9}\cmidrule(lr){10-10}\cmidrule(lr){11-18}\cmidrule(lr){19-19}
 & \shortstack{YOLO\\11x$^\dagger$} & \shortstack{YOLO\\26x} & \shortstack{Faster\\R-CNN} & \shortstack{Retina\\Net} & \shortstack{Cascade\\R-CNN} & \shortstack{RTD\\v2} & \shortstack{D-FINE\\L} & Mean & Mean & \shortstack{YOLO\\11x$^\dagger$} & \shortstack{YOLO\\26x} & \shortstack{Faster\\R-CNN} & \shortstack{Retina\\Net} & \shortstack{Cascade\\R-CNN} & \shortstack{RTD\\v2} & \shortstack{D-FINE\\L} & Mean & Mean \\
\midrule
Visible & 36.59 & 38.31 & 34.80 & 34.22 & 34.23 & 35.88 & 36.86 & 35.84 & 64.73 & 39.88 & 40.15 & 33.40 & 29.45 & 30.34 & 37.40 & 38.51 & 35.59 & 68.88 \\
Infrared & 32.70 & 31.11 & 30.05 & 28.45 & 26.77 & 29.76 & 31.12 & 29.99 & 65.74 & 47.68 & 47.80 & 38.62 & 33.20 & 35.27 & 43.85 & 46.67 & 41.87 & 68.06 \\
\midrule
Text-IF~\cite{yi2024textif} & 35.65 & 34.69 & 31.25 & 28.34 & 29.20 & 32.22 & 34.27 & 32.23 & 65.36 & 51.01 & 51.76 & 41.95 & 36.83 & 35.90 & 44.58 & \textcolor[RGB]{0,0,220}{\underline{48.31}} & 44.33 & 73.32 \\
OmniFuse~\cite{zhang2025omnifuse} & 35.89 & 35.53 & 30.21 & 28.68 & 29.01 & 32.83 & 32.95 & 32.16 & 63.33 & 48.93 & 50.74 & 39.15 & 36.26 & 34.53 & 44.62 & 46.76 & 43.00 & 72.87 \\
HCLFuse~\cite{guo2025hclfuse} & 34.52 & 33.64 & 31.80 & 31.39 & 29.70 & 31.59 & 33.05 & 32.24 & 65.10 & 49.86 & 50.96 & 40.44 & 37.73 & \textcolor[RGB]{0,0,220}{\underline{38.72}} & 44.33 & 46.27 & 44.04 & 72.48 \\
DSPFusion~\cite{tang2026dspfusion} & 35.51 & 34.15 & 31.19 & 31.30 & 30.43 & 33.53 & 33.68 & 32.83 & \textcolor[RGB]{0,0,220}{\underline{66.63}} & \textcolor[RGB]{0,0,220}{\underline{51.54}} & \textcolor[RGB]{0,0,220}{\underline{52.27}} & \textcolor[RGB]{0,0,220}{\underline{43.51}} & \textcolor[RGB]{0,0,220}{\underline{39.94}} & 37.91 & \textcolor[RGB]{0,0,220}{\underline{46.41}} & 48.21 & \textcolor[RGB]{0,0,220}{\underline{45.68}} & 73.88 \\
DuS-DiFuse~\cite{cao2026dusdifuse} & 34.74 & 35.25 & 30.86 & 27.72 & 28.65 & 33.37 & 35.18 & 32.25 & 63.83 & 50.75 & 50.37 & 43.31 & 37.23 & 36.02 & 45.19 & 47.57 & 44.35 & 73.52 \\
DiTFuse~\cite{li2026ditfuse} & 30.00 & 27.53 & 24.54 & 20.96 & 20.80 & 25.02 & 26.85 & 25.10 & 56.19 & 39.60 & 39.79 & 33.33 & 29.10 & 29.97 & 35.93 & 37.93 & 35.09 & 68.21 \\
ISFuse~\cite{wang2026isfuse} & 34.16 & 32.88 & 30.54 & 29.50 & 29.21 & 31.40 & 32.54 & 31.46 & 65.58 & 50.42 & 50.12 & 42.73 & 38.82 & 38.29 & 45.10 & 46.79 & 44.61 & \textcolor[RGB]{0,0,220}{\underline{74.11}} \\
\midrule
Retinexformer~\cite{cai2023retinexformer} & 35.69 & 35.72 & 33.59 & 30.99 & 32.46 & 34.84 & 34.90 & 34.03 & 61.28 & 34.65 & 35.83 & 29.26 & 24.77 & 25.95 & 34.60 & 37.03 & 31.73 & 62.03 \\
DiffLL~\cite{jiang2023diffll} & 25.75 & 23.89 & 31.64 & 27.98 & 31.57 & 33.36 & 34.56 & 29.82 & 54.39 & 31.56 & 28.30 & 27.60 & 23.46 & 24.25 & 35.46 & 37.45 & 29.72 & 58.72 \\
QuadPrior~\cite{wang2024quadprior} & 36.04 & 37.18 & 34.20 & 32.50 & 33.17 & 34.97 & 36.08 & 34.87 & 63.17 & 38.41 & 40.21 & 33.16 & 30.31 & 30.29 & 36.36 & 38.45 & 35.31 & 69.08 \\
DarkIR~\cite{feijoo2025darkir} & \textcolor[RGB]{0,0,220}{\underline{36.67}} & \textcolor[RGB]{0,0,220}{\underline{37.88}} & \textcolor[RGB]{0,0,220}{\underline{34.34}} & \textcolor[RGB]{0,0,220}{\underline{33.85}} & \textcolor[RGB]{0,0,220}{\underline{34.50}} & 36.04 & \textcolor[RGB]{0,0,220}{\underline{36.99}} & \textcolor[RGB]{0,0,220}{\underline{35.75}} & 65.05 & 37.65 & 38.54 & 31.74 & 27.53 & 27.51 & 36.40 & 38.11 & 33.92 & 65.94 \\
HVI-CIDNet~\cite{yan2025hvi} & 35.29 & 37.23 & 32.21 & 31.82 & 33.33 & 35.78 & 35.51 & 34.45 & 62.36 & 34.00 & 35.42 & 27.79 & 22.32 & 24.86 & 33.88 & 36.62 & 30.70 & 60.09 \\
Diff-Retinex++~\cite{yi2025diffretinexpp} & 35.55 & 35.95 & 33.76 & 32.67 & 33.99 & \textcolor[RGB]{0,0,220}{\underline{36.62}} & 36.63 & 35.02 & 63.21 & 34.55 & 35.42 & 29.41 & 25.39 & 25.90 & 34.54 & 38.21 & 31.92 & 62.73 \\
MagicFuse~\cite{zhang2026magicfuse} & 31.84 & 31.73 & 27.71 & 23.36 & 25.93 & 29.99 & 30.45 & 28.71 & 54.46 & 36.81 & 37.85 & 27.73 & 23.03 & 23.93 & 33.37 & 35.91 & 31.24 & 59.39 \\
\midrule
\textbf{NatureFusion} & \textcolor[RGB]{220,0,0}{\textbf{42.44}} & \textcolor[RGB]{220,0,0}{\textbf{42.48}} & \textcolor[RGB]{220,0,0}{\textbf{39.82}} & \textcolor[RGB]{220,0,0}{\textbf{40.69}} & \textcolor[RGB]{220,0,0}{\textbf{41.51}} & \textcolor[RGB]{220,0,0}{\textbf{43.50}} & \textcolor[RGB]{220,0,0}{\textbf{43.18}} & \textcolor[RGB]{220,0,0}{\textbf{41.95}} & \textcolor[RGB]{220,0,0}{\textbf{74.57}} & \textcolor[RGB]{220,0,0}{\textbf{53.05}} & \textcolor[RGB]{220,0,0}{\textbf{52.44}} & \textcolor[RGB]{220,0,0}{\textbf{49.06}} & \textcolor[RGB]{220,0,0}{\textbf{49.86}} & \textcolor[RGB]{220,0,0}{\textbf{49.18}} & \textcolor[RGB]{220,0,0}{\textbf{50.48}} & \textcolor[RGB]{220,0,0}{\textbf{51.50}} & \textcolor[RGB]{220,0,0}{\textbf{50.79}} & \textcolor[RGB]{220,0,0}{\textbf{86.91}} \\
\bottomrule
\end{tabular*}
\endgroup
\end{minipage}}
\end{table*}

Fig.~\ref{fig:qualitative_detection} shows results from the same frozen RetinaNet. Text-IF, OmniFuse, DuS-DiFuse, and DiTFuse fail to detect the pedestrian in the red circles. Among the seven single-modality enhancement methods, only QuadPrior detects this target. NatureFusion renders a discernible dark upper garment and legs in Fig.~\ref{fig:qualitative_detection}(p), with a detection confidence of 0.98, compared with 0.35 on the original visible image and 0.79 on the infrared image. Ten of the fourteen compared methods miss this target, indicating that increasing brightness or directly presenting infrared contours does not necessarily lead to effective target recognition by the detector.

Table~\ref{tab:detection_comparison} reports $\mathrm{AP}_{50:95}$ for each detector and $\mathrm{AP}_{50:95}$ and $\mathrm{AP}_{50}$ averaged across detectors. NatureFusion ranks first for every detector on both sets. The results show that the performance of the compared methods depends on which input the detectors handle better. On SMOD, detectors perform worse on infrared than on visible images, and all seven multi-modal fusion methods also fall below the original visible input, making the single-modality enhancement method DarkIR the strongest competitor. On LLVIP, where infrared is the stronger input, most fusion methods instead outperform enhancement methods. This pattern is consistent with the source-fidelity paradigm, as outputs that retain infrared appearance inherit both its benefits and its drawbacks for detectors pretrained on visible images. NatureFusion exceeds both original inputs on both sets. Moreover, its advantage is not concentrated on YOLO11x, which provides task adaptation supervision. On LLVIP, NatureFusion leads by about 1.5 percentage points with YOLO11x but about 10 percentage points with RetinaNet and Cascade R-CNN, showing that its performance gains transfer to detection architectures excluded from training.

\begin{figure*}[!t]
\centering
\includegraphics[width=\textwidth]{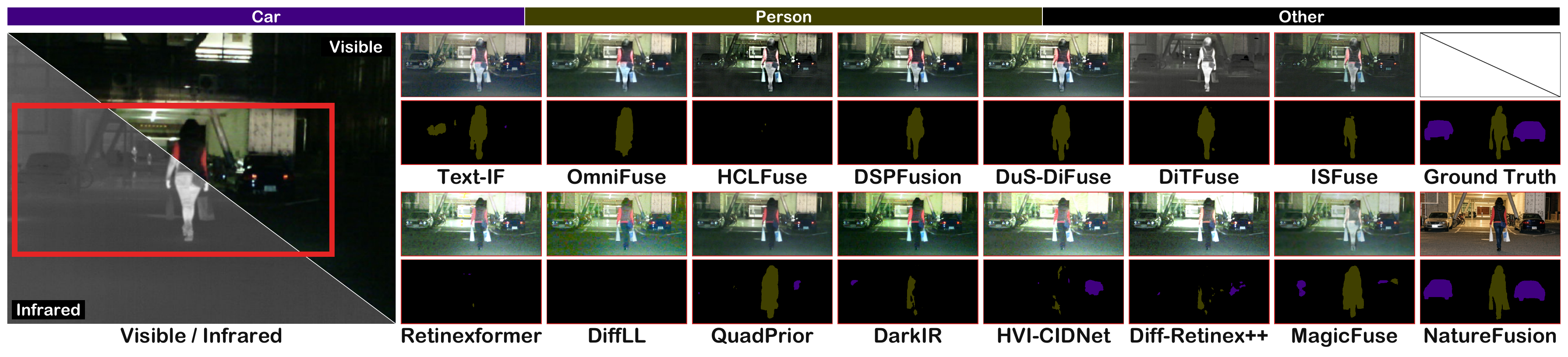}
\caption{Frozen segmentation on MFNet by OCRNet-HR48. The left panel shows the visible and infrared inputs, with the red box marking the enlarged region, and each method's crop is shown above its prediction.}
\label{fig:qualitative_segmentation}
\par\vspace{4pt}
\end{figure*}

Figs.~\ref{fig:det_iou_profile}(a) and (b) show mean AP over the seven detectors as a function of the IoU threshold. NatureFusion outperforms all fourteen compared methods and both original inputs over thresholds from 0.30 to 0.85 on SMOD and from 0.30 to 0.80 on LLVIP, with different patterns in its margin on the two sets. On LLVIP, the margin is largest at loose thresholds, reaching about 14 percentage points at 0.30 and remaining at 3.4 percentage points at 0.75. This indicates that much of the improvement comes from recovering pedestrians missed in visible images, consistent with the 47.1\% recall of such pedestrians on LLVIP in Table~\ref{tab:lle_detection_comparison}. On SMOD, the margin instead grows with the threshold, from about 7 percentage points at 0.30 to about 12 at 0.65, and remains at 7.6 at 0.75. This shows that the advantage lies in both detecting more targets and localizing them more precisely. This agrees with Fig.~\ref{fig:qualitative_fusion}, where large areas of clothing and body parts are washed out in the compared outputs, while NatureFusion renders clothing and limbs with clear boundaries, allowing detection boxes to fit more tightly. At the strictest thresholds, AP drops sharply for all methods and the curves converge, with differences within 2 percentage points.

\begin{table*}[!t]
\centering
\makebox[\linewidth][c]{\begin{minipage}{\textwidth}
\centering
\caption{Frozen segmentation comparison on MFNet (\%).}
\label{tab:segmentation_comparison}
\begingroup
\fontsize{8}{10}\selectfont
\setlength{\tabcolsep}{1.5pt}
\renewcommand{\arraystretch}{1.1}
\begin{tabular*}{\textwidth}{@{\extracolsep{\fill}}lccccccccccccccccc@{}}
\toprule
\multirow{2}{*}[-1.5pt]{Method} & \multicolumn{2}{c}{PSPNet} & \multicolumn{2}{c}{DLV3+} & \multicolumn{2}{c}{OCRNet} & \multicolumn{2}{c}{SegFormer} & \multicolumn{2}{c}{Mask2F} & \multicolumn{2}{c}{SegMAN} & \multicolumn{2}{c}{Y26-sem} & \multicolumn{3}{c}{Mean} \\
\cmidrule(lr){2-3}\cmidrule(lr){4-5}\cmidrule(lr){6-7}\cmidrule(lr){8-9}\cmidrule(lr){10-11}\cmidrule(lr){12-13}\cmidrule(lr){14-15}\cmidrule(lr){16-18}
 & Car & Person & Car & Person & Car & Person & Car & Person & Car & Person & Car & Person & Car & Person & Car & Person & $\mathrm{mIoU}_2$ \\
\midrule
Visible & 34.01 & 25.11 & 36.12 & 27.12 & 41.52 & 30.95 & 59.14 & 39.26 & 48.62 & 50.56 & 59.12 & 54.23 & 55.78 & 45.46 & 47.76 & 38.96 & 43.36 \\
Infrared & 13.67 & 25.03 & 16.44 & 41.14 & 12.51 & 36.07 & 27.97 & 45.56 & 31.63 & 57.21 & 48.81 & 58.31 & 31.94 & 66.47 & 26.14 & 47.11 & 36.63 \\
\midrule
Text-IF~\cite{yi2024textif} & 37.14 & \textcolor[RGB]{0,0,220}{\underline{45.19}} & \textcolor[RGB]{0,0,220}{\underline{37.26}} & 47.98 & \textcolor[RGB]{0,0,220}{\underline{42.69}} & 55.82 & 58.89 & 55.05 & 45.69 & 59.41 & 62.12 & 61.62 & 54.19 & 62.38 & \textcolor[RGB]{0,0,220}{\underline{48.28}} & 55.35 & \textcolor[RGB]{0,0,220}{\underline{51.82}} \\
OmniFuse~\cite{zhang2025omnifuse} & 32.07 & 26.79 & 27.60 & 23.65 & 33.10 & 39.79 & 53.70 & 45.20 & 44.00 & 53.49 & 53.22 & 56.28 & 50.46 & 54.81 & 42.02 & 42.86 & 42.44 \\
HCLFuse~\cite{guo2025hclfuse} & 6.89 & 19.05 & 13.47 & 34.89 & 23.02 & 45.74 & 47.04 & 57.22 & 35.75 & 63.44 & 55.13 & 61.19 & 44.59 & 63.32 & 32.27 & 49.27 & 40.77 \\
DSPFusion~\cite{tang2026dspfusion} & 30.75 & 34.38 & 32.75 & 41.52 & 39.03 & 55.60 & 58.49 & 58.63 & 48.02 & 60.21 & 61.93 & 63.28 & 54.23 & 62.86 & 46.46 & 53.78 & 50.12 \\
DuS-DiFuse~\cite{cao2026dusdifuse} & 31.99 & 44.29 & 32.52 & 49.98 & 36.51 & \textcolor[RGB]{0,0,220}{\underline{59.33}} & \textcolor[RGB]{0,0,220}{\underline{59.73}} & \textcolor[RGB]{0,0,220}{\underline{61.55}} & 41.74 & 62.05 & \textcolor[RGB]{0,0,220}{\underline{62.48}} & \textcolor[RGB]{0,0,220}{\underline{64.29}} & 52.37 & 63.21 & 45.34 & \textcolor[RGB]{0,0,220}{\underline{57.82}} & 51.58 \\
DiTFuse~\cite{li2026ditfuse} & 21.99 & 31.16 & 24.88 & 44.09 & 22.92 & 46.90 & 36.48 & 45.74 & 31.57 & 58.35 & 38.37 & 55.46 & 37.51 & 62.73 & 30.53 & 49.20 & 39.87 \\
ISFuse~\cite{wang2026isfuse} & 21.71 & 43.40 & 25.49 & \textcolor[RGB]{0,0,220}{\underline{54.08}} & 30.72 & 54.88 & 52.47 & 56.61 & 45.84 & \textcolor[RGB]{0,0,220}{\underline{64.00}} & 57.26 & 62.11 & 46.72 & \textcolor[RGB]{0,0,220}{\underline{65.93}} & 40.03 & 57.29 & 48.66 \\
\midrule
Retinexformer~\cite{cai2023retinexformer} & 22.46 & 19.56 & 18.35 & 19.53 & 29.49 & 24.13 & 51.98 & 30.62 & 39.19 & 39.34 & 57.19 & 45.02 & 45.81 & 35.77 & 37.78 & 30.57 & 34.17 \\
DiffLL~\cite{jiang2023diffll} & 13.16 & 12.52 & 21.02 & 14.49 & 23.00 & 18.12 & 45.26 & 24.45 & 33.52 & 32.40 & 59.28 & 46.04 & 32.34 & 24.22 & 32.51 & 24.61 & 28.56 \\
QuadPrior~\cite{wang2024quadprior} & \textcolor[RGB]{0,0,220}{\underline{37.39}} & 30.09 & 36.38 & 30.43 & 41.06 & 36.80 & 58.05 & 46.67 & 41.32 & 48.22 & 60.08 & 53.48 & \textcolor[RGB]{0,0,220}{\underline{54.87}} & 49.31 & 47.02 & 42.14 & 44.58 \\
DarkIR~\cite{feijoo2025darkir} & 18.47 & 19.76 & 18.97 & 17.19 & 26.16 & 26.50 & 56.22 & 41.70 & 38.33 & 41.99 & 57.11 & 47.80 & 48.78 & 36.84 & 37.72 & 33.11 & 35.42 \\
HVI-CIDNet~\cite{yan2025hvi} & 24.64 & 18.49 & 23.46 & 17.59 & 32.36 & 26.57 & 54.40 & 34.63 & 38.73 & 43.80 & 59.95 & 50.77 & 47.13 & 37.58 & 40.09 & 32.78 & 36.44 \\
Diff-Retinex++~\cite{yi2025diffretinexpp} & 34.49 & 25.09 & 31.12 & 24.56 & 39.75 & 27.30 & 53.97 & 33.05 & 41.28 & 37.41 & 59.64 & 50.08 & 50.10 & 40.08 & 44.34 & 33.94 & 39.14 \\
MagicFuse~\cite{zhang2026magicfuse} & 29.15 & 17.54 & 26.14 & 15.19 & 40.93 & 39.78 & 58.12 & 41.71 & \textcolor[RGB]{0,0,220}{\underline{51.40}} & 51.58 & 58.95 & 46.83 & 52.84 & 50.95 & 45.36 & 37.65 & 41.51 \\
\midrule
\textbf{NatureFusion} & \textcolor[RGB]{220,0,0}{\textbf{69.08}} & \textcolor[RGB]{220,0,0}{\textbf{66.15}} & \textcolor[RGB]{220,0,0}{\textbf{64.98}} & \textcolor[RGB]{220,0,0}{\textbf{66.81}} & \textcolor[RGB]{220,0,0}{\textbf{67.32}} & \textcolor[RGB]{220,0,0}{\textbf{65.55}} & \textcolor[RGB]{220,0,0}{\textbf{67.88}} & \textcolor[RGB]{220,0,0}{\textbf{67.34}} & \textcolor[RGB]{220,0,0}{\textbf{69.23}} & \textcolor[RGB]{220,0,0}{\textbf{68.64}} & \textcolor[RGB]{220,0,0}{\textbf{67.61}} & \textcolor[RGB]{220,0,0}{\textbf{67.96}} & \textcolor[RGB]{220,0,0}{\textbf{70.66}} & \textcolor[RGB]{220,0,0}{\textbf{66.10}} & \textcolor[RGB]{220,0,0}{\textbf{68.11}} & \textcolor[RGB]{220,0,0}{\textbf{66.94}} & \textcolor[RGB]{220,0,0}{\textbf{67.52}} \\
\bottomrule
\end{tabular*}
\endgroup
\end{minipage}}
\end{table*}

\begin{table}[!t]
\centering
\begin{minipage}{\columnwidth}
\centering
\caption{Retrained segmentation on MFNet.}
\label{tab:segmentation_retrain_comparison}
\begingroup
\fontsize{8}{10}\selectfont
\setlength{\tabcolsep}{1.5pt}
\renewcommand{\arraystretch}{1.08}
\resizebox{\columnwidth}{!}{\begin{tabular}{@{}lcccccccc@{}}
\toprule
\multirow{2}{*}[-1.2pt]{Method} & \multicolumn{4}{c}{$\mathrm{mIoU}_{9}\uparrow$} & \multicolumn{4}{c}{$\mathrm{mIoU}_{2}\uparrow$} \\
\cmidrule(lr){2-5}\cmidrule(l){6-9}
 & PSPNet & DLV3+ & SegFormer & Mean & PSPNet & DLV3+ & SegFormer & Mean \\
\midrule
Visible & 46.07 & 45.82 & 48.48 & 46.79 & 62.12 & 62.75 & 65.34 & 63.41 \\
Infrared & 42.18 & 39.74 & 43.98 & 41.97 & 70.00 & 70.83 & 71.48 & 70.77 \\
\midrule
Text-IF~\cite{yi2024textif} & \textcolor[RGB]{0,0,220}{\underline{49.78}} & 45.88 & 49.96 & \textcolor[RGB]{0,0,220}{\underline{48.54}} & \textcolor[RGB]{0,0,220}{\underline{72.42}} & \textcolor[RGB]{0,0,220}{\underline{73.47}} & 74.02 & \textcolor[RGB]{0,0,220}{\underline{73.30}} \\
OmniFuse~\cite{zhang2025omnifuse} & 45.75 & 46.77 & 48.88 & 47.13 & 65.91 & 64.98 & 70.79 & 67.23 \\
HCLFuse~\cite{guo2025hclfuse} & 47.49 & 42.72 & 47.93 & 46.04 & 66.73 & 67.22 & 71.47 & 68.47 \\
DSPFusion~\cite{tang2026dspfusion} & 47.94 & 43.75 & \textcolor[RGB]{0,0,220}{\underline{51.10}} & 47.60 & 70.11 & 73.19 & 74.11 & 72.47 \\
DuS-DiFuse~\cite{cao2026dusdifuse} & 47.58 & 43.70 & 50.76 & 47.34 & 71.97 & \textcolor[RGB]{0,0,220}{\underline{73.47}} & \textcolor[RGB]{0,0,220}{\underline{74.30}} & 73.24 \\
DiTFuse~\cite{li2026ditfuse} & 39.07 & 39.81 & 42.51 & 40.46 & 66.77 & 68.39 & 70.42 & 68.53 \\
ISFuse~\cite{wang2026isfuse} & 47.24 & 44.66 & 47.86 & 46.59 & 68.47 & 70.23 & 71.72 & 70.14 \\
\midrule
HVI-CIDNet~\cite{yan2025hvi} & 43.79 & 40.33 & 46.05 & 43.39 & 61.25 & 60.25 & 62.42 & 61.30 \\
MagicFuse~\cite{zhang2026magicfuse} & 46.04 & \textcolor[RGB]{0,0,220}{\underline{46.88}} & 48.52 & 47.15 & 64.55 & 66.91 & 66.80 & 66.09 \\
\midrule
\textbf{NatureFusion} & \textcolor[RGB]{220,0,0}{\textbf{52.86}} & \textcolor[RGB]{220,0,0}{\textbf{48.89}} & \textcolor[RGB]{220,0,0}{\textbf{53.76}} & \textcolor[RGB]{220,0,0}{\textbf{51.83}} & \textcolor[RGB]{220,0,0}{\textbf{73.83}} & \textcolor[RGB]{220,0,0}{\textbf{75.03}} & \textcolor[RGB]{220,0,0}{\textbf{76.13}} & \textcolor[RGB]{220,0,0}{\textbf{75.00}} \\
\bottomrule
\end{tabular}}
\endgroup
\end{minipage}
\end{table}

\looseness=-1
\textbf{Frozen segmentation.} On MFNet-250, the seven Cityscapes-pretrained segmenters listed in Sec.~\ref{subsec:experimental_setup} are used to evaluate Car IoU, Person IoU, and their mean $\mathrm{mIoU}_2$. Fig.~\ref{fig:qualitative_segmentation} shows results from the frozen OCRNet-HR48. Text-IF, OmniFuse, DSPFusion, and DuS-DiFuse segment the main pedestrian but incompletely predict the vehicles on either side. QuadPrior and MagicFuse also retain the pedestrian region, whereas their vehicle predictions are fragmented. NatureFusion produces more complete predictions for both vehicles and the pedestrian, with the main spatial layout consistent with the annotations. This illustrates the joint representation of infrared-salient targets and visible scene structure in its fused images.

Table~\ref{tab:segmentation_comparison} reports quantitative results with frozen segmenters. The two source images show clear complementarity across target classes. Visible inputs yield substantially better car segmentation than infrared inputs, whereas infrared inputs provide more salient pedestrian cues and perform better on person segmentation. However, existing methods often struggle to accommodate both classes. Text-IF obtains Car IoU close to the visible input, and DuS-DiFuse and ISFuse perform well on person segmentation, but HCLFuse and DiTFuse show substantial drops in Car IoU. In contrast, NatureFusion ranks first in all fourteen evaluations across seven segmenters. Its mean Car and Person IoU reach 68.11 and 66.94, exceeding the better source input for each class by 20.35 and 19.83 percentage points, respectively. NatureFusion also performs consistently across segmenters, whereas some compared methods fluctuate considerably. For example, HCLFuse obtains a Car IoU of only 6.89 with PSPNet. These results show that NatureFusion effectively uses complementary visible and infrared information to improve segmentation of both cars and persons without sacrificing either class.

\textbf{Retrained segmentation.} To assess performance after segmenters adapt to each method's outputs, we separately retrain PSPNet-R50, DeepLabV3+ R50, and SegFormer-B2. The comparison includes the seven multi-modal fusion methods, HVI-CIDNet, MagicFuse, NatureFusion, and both original inputs. All input representations share the same 100 training samples and are evaluated on 191 MFNet samples acquired on different dates from the training samples. For each segmenter, all methods start from identical Cityscapes-pretrained weights, with the classifier replaced by a newly initialized nine-class layer. We use a batch size of 8, 1000 training iterations, and a random seed of 42, and evaluate the model from the final iteration. Table~\ref{tab:segmentation_retrain_comparison} reports $\mathrm{mIoU}_9$ over all nine classes and $\mathrm{mIoU}_2$ over Car and Person.

NatureFusion achieves the best performance on both metrics for all three segmenters, with mean $\mathrm{mIoU}_9$ and $\mathrm{mIoU}_2$ of 51.83 and 75.00, exceeding the respective second-best methods by 3.29 and 1.70 percentage points. Compared with frozen-segmenter evaluation, its margins narrow after retraining, and all compared fusion methods exceed the visible input in $\mathrm{mIoU}_2$. This indicates that retraining helps segmenters adapt to fused images and mitigates performance losses caused by differences in input distributions. Nevertheless, NatureFusion retains its lead on all metrics, showing that its advantage comes not only from more natural visible appearance but also from exploiting complementary information from both modalities to provide more useful target cues for segmentation.

\begin{table}[!t]
\centering
\begin{minipage}{\columnwidth}
\centering
\caption{Role of infrared at each stage on LLVIP. w/o IR replaces infrared with an all-zero image at inference only, and VI+VI replaces infrared with the visible image in both training and inference. The best value in each column is marked in bold red.}

\label{tab:ir_necessity}
\begingroup
\fontsize{8}{9.6}\selectfont
\setlength{\tabcolsep}{0.7pt}
\renewcommand{\arraystretch}{1.24}
\begin{tabular*}{\columnwidth}{@{\extracolsep{\fill}}lcccccc@{}}
\toprule
Stages & Setting & $\mathrm{AP}_{50:95}\uparrow$ & $\mathrm{AP}_{50}\uparrow$ & $\mathrm{AP}_{75}\uparrow$ & $\mathrm{AP}_{90}\uparrow$ & $R_{\mathrm{miss}}\uparrow$ \\
\midrule
Test & NatureFusion, w/o IR & 32.33 & 63.69 & 32.20 & 2.23 & 23.2 \\
\midrule
\multirow{2}{*}{I} & Teacher w/o IR & 25.99 & 58.46 & 22.21 & 1.62 & 21.6 \\
 & Teacher w/ IR & 25.69 & 58.19 & 20.64 & 1.20 & 29.8 \\
\midrule
\multirow{2}{*}{II} & Fusion-only, VI+VI & 32.12 & 63.62 & 31.89 & 2.42 & 24.1 \\
 & Fusion-only, VI+IR & 34.87 & 67.57 & 34.85 & 3.03 & 29.0 \\
\midrule
\multirow{2}{*}{II+III} & Fusion + Task, VI+VI & 32.90 & 65.04 & 32.18 & 1.81 & 25.1 \\
 & \textbf{NatureFusion, VI+IR} & \textcolor[RGB]{220,0,0}{\textbf{35.61}} & \textcolor[RGB]{220,0,0}{\textbf{68.59}} & \textcolor[RGB]{220,0,0}{\textbf{35.42}} & \textcolor[RGB]{220,0,0}{\textbf{3.23}} & \textcolor[RGB]{220,0,0}{\textbf{33.9}} \\
\bottomrule
\end{tabular*}
\endgroup
\end{minipage}
\end{table}

\subsection{Ablation Studies}
\label{subsec:ablation}

This section analyzes the roles of infrared input, fusion learning, and task adaptation through ablation studies and examines the amplification factor $w$ for the complementary infrared response. Component ablations are conducted on the LLVIP-NCE test split, using missed pedestrian recall $R_{\mathrm{miss}}$ and the no-reference quality metric NIQE as defined in Sec.~\ref{subsec:experimental_setup}.

\textbf{Role of infrared at each stage.} Table~\ref{tab:ir_necessity} analyzes the role of infrared information in teacher demonstration generation, student fusion learning, and model inference. Both teacher configurations in Stage I use person bounding boxes, and their demonstrations are regenerated for this comparison. Adding infrared raises the teacher's $R_{\mathrm{miss}}$ from 21.6\% to 29.8\%, but none of the AP metrics improves, and $\mathrm{AP}_{75}$ even decreases. This indicates that infrared guidance helps the teacher depict more pedestrians that are difficult to discern, although the generated targets may still exhibit positional errors, consistent with the observations in Sec.~\ref{subsec:further_analysis}. The student uses infrared information more effectively. With the same teacher demonstrations, replacing the second visible input with infrared improves all four AP metrics and $R_{\mathrm{miss}}$, indicating that the student uses target location cues from infrared rather than relying entirely on the demonstrations. Adding task adaptation further raises $\mathrm{AP}_{50:95}$ and $R_{\mathrm{miss}}$ to 35.61 and 33.9\%, respectively. Removing infrared only at inference reduces them to 32.33 and 23.2\%, approaching the configurations trained with visible inputs alone. These results show that NatureFusion's gains come not only from its learned prior for natural image generation but also from effective use of complementary infrared information during inference.

\begin{figure}[!t]
    \centering
    \includegraphics[width=3.0in]{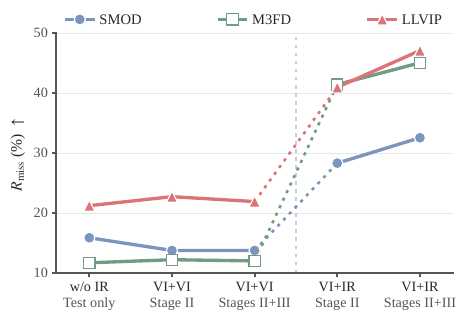}
    \caption{$R_{\mathrm{miss}}$ under different infrared settings.}
    \label{fig:ir_dependence}
\par\vspace{\floatsep}
    \centering
    \includegraphics[width=\columnwidth]{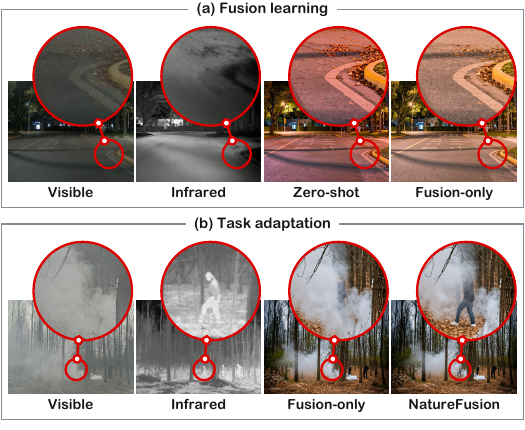}
    \caption{Qualitative ablation of (a) fusion learning on SMOD and (b) task adaptation on M3FD.}
    \label{fig:component_ablation}
\end{figure}

Fig.~\ref{fig:ir_dependence} further compares missed pedestrian recall on SMOD-200, M3FD-150, and LLVIP-250. To evaluate the contribution of infrared, the infrared-free control uses two copies of the visible image during both training and inference, sets $w=1$, and disables the anchoring loss in Stage III. With fusion learning alone, adding infrared raises $R_{\mathrm{miss}}$ to 28.3\% to 41.4\%, approximately twice the 12.2\% to 22.7\% obtained with duplicated visible inputs. Task adaptation adds a further 3.6 to 6.1 percentage points when infrared is available, while the infrared-free configuration obtains no corresponding gain. This indicates that task adaptation further exploits complementary infrared information. Together with the results of removing infrared only at test time (w/o~IR), these comparisons show that using infrared during both training and inference improves the discernibility of pedestrians missed in visible images.

\textbf{Fusion learning.} Table~\ref{tab:component_ablation} analyzes the roles of fusion learning and task adaptation through six configurations. First, the zero-shot backbone and Fusion-only use identical visible and infrared image pairs and text prompts, with the latter additionally loading the fusion adapter trained on teacher demonstrations. Adding the fusion adapter reduces NIQE from 3.216 to 2.774 and visibly alleviates the reddish road in the zero-shot output in Fig.~\ref{fig:component_ablation}(a). Detection performance changes little, with $R_{\mathrm{miss}}$ rising slightly from 28.0\% to 29.2\% and $\mathrm{AP}_{50:95}$ decreasing slightly from 34.62 to 34.44. These results indicate that the fusion adapter primarily improves natural appearance by learning from teacher demonstrations, while further detection gains depend on task adaptation.

\textbf{Task adaptation.} Introducing the task adapter while keeping the fusion adapter frozen further improves all metrics, with a particularly pronounced gain in missed pedestrian recall. Compared with Fusion-only, $R_{\mathrm{miss}}$ increases by 3.5 percentage points, $\mathrm{AP}_{50:95}$ increases by 0.83 percentage points, and NIQE decreases from 2.774 to 2.738. This indicates that task adaptation further exploits the complementary contribution of infrared for targets that are difficult to discern in visible images. As shown in Fig.~\ref{fig:component_ablation}(b), the infrared image reveals the contour of a person within the smoke, but Fusion-only still struggles to render the body and legs clearly. After task adaptation, NatureFusion produces a clearer body contour and distinctly separated legs. The qualitative and quantitative results together show that task adaptation further improves target \mbox{discernibility beyond fusion learning.}

\looseness=1
To further analyze the individual designs in task adaptation, we compare three ablation configurations. With $w=1$, the complementarity loss remains active, but its supervision target equals the Fusion-only prediction without amplifying the complementary infrared response. In this case, $\mathrm{AP}_{50:95}$ increases from 34.44 to 34.85, whereas $R_{\mathrm{miss}}$ remains at 29.2\%, indicating that amplification of the complementary infrared response is particularly important for improving missed pedestrian recall. Removing the detection loss or the anchoring loss reduces $R_{\mathrm{miss}}$ to 31.4\% and 31.3\%, respectively, and both configurations also have lower $\mathrm{AP}_{50:95}$ than the full model. These results show that amplification of the complementary infrared response, detection supervision, and anchoring work together to \mbox{further improve target discernibility.}

\textbf{Amplification factor.} Table~\ref{tab:w_sensitivity} compares the effects of different amplification factors $w$ on M3FD-150. As $w$ increases from 1 to 2, overall detection performance generally improves, with all five metrics reaching their best values at $w=2$. The largest gain occurs in $\mathrm{AP}_{70}$, which increases by 2.64 percentage points over $w=1$, while improvements in $\mathrm{AP}_{90}$ and $\mathrm{AP}_{95}$ are more limited. This indicates that amplifying the complementary infrared response mainly improves detection at moderate IoU thresholds, with smaller gains in precise target localization. Increasing $w$ further to 3 brings no additional improvement in any metric, showing that excessive amplification of the infrared response provides no further benefit. Therefore, we adopt $w=2$.

\begin{table}[!t]
\centering
\begin{minipage}{\columnwidth}
\centering
\caption{Component ablation on LLVIP. Zero-shot is the Distilled backbone without adapters, and other variants report the mean of two training runs.}

\label{tab:component_ablation}
\fontsize{8}{9.6}\selectfont
\setlength{\tabcolsep}{1.5pt}
\renewcommand{\arraystretch}{1.24}
\begin{tabular*}{\columnwidth}{@{\extracolsep{\fill}}lccccc@{}}
\toprule
\multirow{2}{*}[-1.2pt]{Variant} & \multicolumn{5}{c}{LLVIP} \\
\cmidrule(lr){2-6}
 & $\mathrm{AP}_{50:95}\uparrow$ & $\mathrm{AP}_{50}\uparrow$ & $\mathrm{AP}_{75}\uparrow$ & $R_{\mathrm{miss}}\uparrow$ & NIQE$\downarrow$ \\
\midrule
Zero-shot (VI+IR) & 34.62 & 67.22 & 33.17 & 28.0 & 3.216 \\
\midrule
Fusion-only & 34.44 & 67.30 & 33.87 & 29.2 & 2.774 \\
$w=1$ & 34.85 & 67.36 & 34.65 & 29.2 & 2.763 \\
w/o $\mathcal{L}_{\mathrm{det}}$ & 34.82 & 67.75 & 34.16 & 31.4 & 2.747 \\
w/o $\mathcal{L}_{\mathrm{anc}}$ & 35.02 & 68.00 & 34.42 & 31.3 & 2.747 \\
\midrule
\textbf{NatureFusion} & \textcolor[RGB]{220,0,0}{\textbf{35.27}} & \textcolor[RGB]{220,0,0}{\textbf{68.26}} & \textcolor[RGB]{220,0,0}{\textbf{34.94}} & \textcolor[RGB]{220,0,0}{\textbf{32.7}} & \textcolor[RGB]{220,0,0}{\textbf{2.738}} \\
\bottomrule
\end{tabular*}
\end{minipage}
\end{table}

\begin{table}[!t]
\centering
\begin{minipage}{\columnwidth}
\centering
\caption{Sensitivity to the amplification factor $w$ on M3FD.}
\label{tab:w_sensitivity}
\fontsize{8}{9.6}\selectfont
\setlength{\tabcolsep}{1.5pt}
\renewcommand{\arraystretch}{1.24}
\begin{tabular*}{\columnwidth}{@{\extracolsep{\fill}}lccccc@{}}
\toprule
$w$ & $\mathrm{AP}_{55}\uparrow$ & $\mathrm{AP}_{70}\uparrow$ & $\mathrm{AP}_{90}\uparrow$ & $\mathrm{AP}_{95}\uparrow$ & $R_{\mathrm{miss}}\uparrow$ \\
\midrule
1 & 77.89 & 65.59 & 11.51 & 1.70 & 43.1 \\
1.5 & 78.00 & 66.21 & 11.90 & 1.63 & 43.4 \\
\textbf{2} & \textcolor[RGB]{220,0,0}{\textbf{78.69}} & \textcolor[RGB]{220,0,0}{\textbf{68.23}} & \textcolor[RGB]{220,0,0}{\textbf{12.01}} & \textcolor[RGB]{220,0,0}{\textbf{1.76}} & \textcolor[RGB]{220,0,0}{\textbf{45.0}} \\
3 & 78.50 & 68.08 & 11.81 & 1.63 & 44.9 \\
\bottomrule
\end{tabular*}
\end{minipage}
\end{table}

\begin{figure}[!t]
    \centering
    \includegraphics[width=\columnwidth]{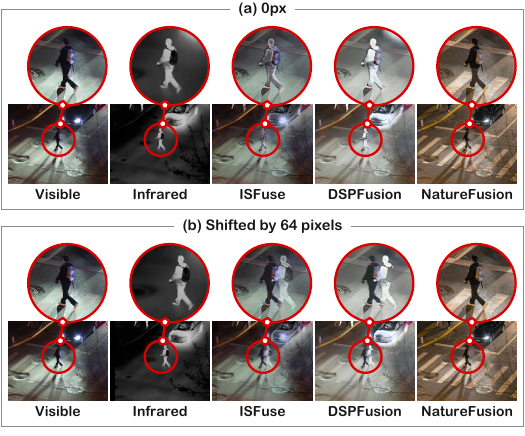}
    \caption{Comparison under registration errors on LLVIP with (a) aligned inputs and (b) infrared shifted right by 64 pixels.}
    \label{fig:misregistration}
\par\vspace{\floatsep}
\begingroup
\captionsetup{type=table}
\centering
\begin{minipage}{\columnwidth}
\centering
\caption{Sensitivity to rightward infrared translation on SMOD.}
\label{tab:misregistration}
\fontsize{8}{10}\selectfont
\setlength{\tabcolsep}{0.5pt}
\renewcommand{\arraystretch}{1.12}
\begin{tabular*}{\columnwidth}{@{\extracolsep{\fill}}lcccccccc@{}}
\toprule
\multirow{2}{*}[-1.5pt]{Method} & \multicolumn{4}{c}{$\mathrm{AP}_{50:95}$ (\%)} & \multicolumn{4}{c}{$R_{\mathrm{miss}}$ (\%)} \\
\cmidrule(lr){2-5}\cmidrule(l){6-9}
 & 0 px & 8 px & 16 px & 32 px & 0 px & 8 px & 16 px & 32 px \\
\midrule
ISFuse & 31.46 & 7.72 & 5.08 & 4.07 & 30.16 & 3.70 & 5.29 & 5.29 \\
DSPFusion & 32.83 & 11.23 & 9.63 & 9.03 & 26.46 & 5.03 & 6.88 & 5.56 \\
NatureFusion & \textcolor[RGB]{220,0,0}{\textbf{41.95}} & \textcolor[RGB]{220,0,0}{\textbf{32.68}} & \textcolor[RGB]{220,0,0}{\textbf{33.19}} & \textcolor[RGB]{220,0,0}{\textbf{36.21}} & \textcolor[RGB]{220,0,0}{\textbf{32.54}} & \textcolor[RGB]{220,0,0}{\textbf{21.69}} & \textcolor[RGB]{220,0,0}{\textbf{22.75}} & \textcolor[RGB]{220,0,0}{\textbf{21.16}} \\
\bottomrule
\end{tabular*}
\end{minipage}
\endgroup

\end{figure}

\begin{figure}[t]
    \centering
    \includegraphics[width=\columnwidth]{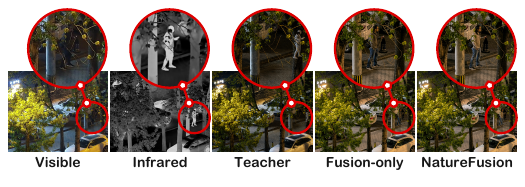}
    \caption{Teacher and student outputs on an LLVIP scene.}
    \label{fig:teacher_student}
\par\vspace{\floatsep}
\begingroup
\captionsetup{type=table}
\centering
\begin{minipage}{\columnwidth}
\centering
\caption{Teacher demonstrations versus student outputs on LLVIP. API indicates whether inference requires an online API.}
\label{tab:teacher_student}
\fontsize{8}{9.6}\selectfont
\setlength{\tabcolsep}{0.3pt}
\renewcommand{\arraystretch}{1.12}
\begin{tabular*}{\columnwidth}{@{\extracolsep{\fill}}lcccccc@{}}
\toprule
\multirow{2}{*}[-1.5pt]{Method} & \multicolumn{2}{c}{Inference} & \multicolumn{3}{c}{Detection} & Quality \\
\cmidrule(lr){2-3}\cmidrule(lr){4-6}\cmidrule(l){7-7}
 & Inputs & API & {\fontsize{7}{8.4}\selectfont $\mathrm{AP}_{50:95}\uparrow$} & {\fontsize{7}{8.4}\selectfont $\mathrm{AP}_{50}\uparrow$} & {\fontsize{7}{8.4}\selectfont $R_{\mathrm{miss}}\uparrow$} & {\fontsize{7}{8.4}\selectfont NIQE$\downarrow$} \\
\midrule
{\fontsize{7.5}{9}\selectfont Visible} & {\fontsize{7.5}{9}\selectfont VI} & No & 23.04 & 48.29 & 0.0 & 3.914 \\
\midrule
{\fontsize{7.5}{9}\selectfont Teacher} & {\fontsize{7.5}{9}\selectfont VI + IR + boxes} & Yes & 25.79 & 55.81 & 26.3 & 3.150 \\
{\fontsize{7.5}{9}\selectfont Fusion-only} & {\fontsize{7.5}{9}\selectfont VI + IR} & No & 34.87 & 67.57 & 29.0 & 2.770 \\
{\fontsize{7.5}{9}\selectfont \textbf{NatureFusion}} & {\fontsize{7.5}{9}\selectfont VI + IR} & No & \textcolor[RGB]{220,0,0}{\textbf{35.61}} & \textcolor[RGB]{220,0,0}{\textbf{68.59}} & \textcolor[RGB]{220,0,0}{\textbf{33.9}} & \textcolor[RGB]{220,0,0}{\textbf{2.718}} \\
\bottomrule
\end{tabular*}
\end{minipage}
\endgroup

\end{figure}

\subsection{Further Analysis}
\label{subsec:further_analysis}

\textbf{Registration errors.} To examine the effect of input misalignment, we fix the model weights and shift the SMOD-200 infrared images to the right by 0, 8, 16, and 32 pixels, padding the exposed area on the left with black. Visible images and annotations remain unchanged. The largest shift corresponds to 5\% of the image width. We compare NatureFusion with ISFuse and DSPFusion, which have the highest $\mathrm{AP}_{50}$ under aligned inputs among the seven compared multi-modal fusion methods. All detection metrics are averaged over the seven frozen detectors.

\begin{figure}[!t]
\centering
    \includegraphics[width=\columnwidth]{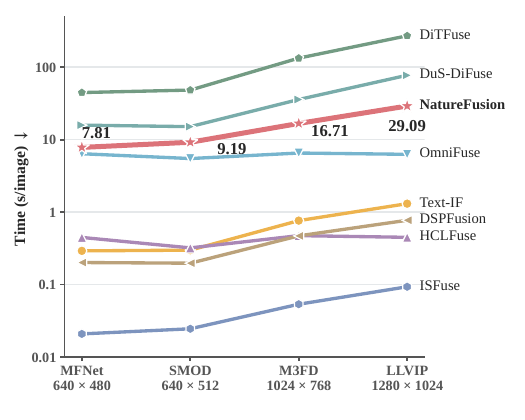}
    \caption{Mean generation time per image at native resolution on a single RTX 3090 (log scale).}
    \label{fig:efficiency}
\par\vspace{\floatsep}
\centering
    \includegraphics[width=\columnwidth]{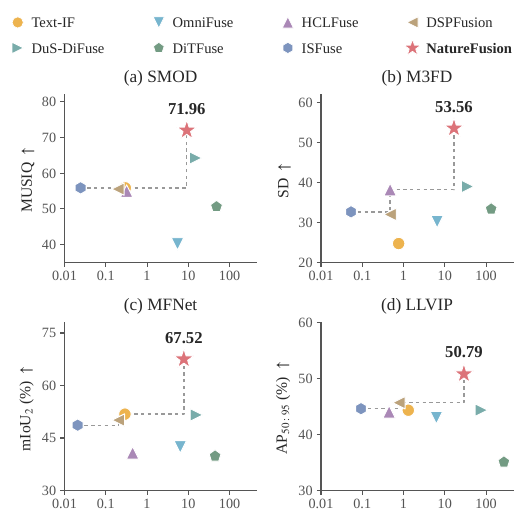}
    \caption{Performance versus generation time (s/image, log scale). Dashed lines trace the best score achievable within each time budget.}
    \label{fig:efficiency_pareto}
\end{figure}

\begingroup
\widowpenalty=10000
Fig.~\ref{fig:misregistration} shows an LLVIP scene in which the visible image already contains clear person contours. The infrared image is shifted 64 pixels to the right. ISFuse and DSPFusion overlay the shifted infrared silhouettes beside the visible pedestrians, producing spatially separated bright contours. In NatureFusion, the pedestrians' body positions, head orientations, and walking poses remain close to those obtained with aligned inputs, rather than shifting together with the infrared contours. This example shows that our method remains stable under input misalignment.\par
\endgroup

Table~\ref{tab:misregistration} reports quantitative results on SMOD-200. The two compared methods are sensitive to infrared misalignment. A shift of only 8 pixels reduces their $\mathrm{AP}_{50:95}$ from above 31 to below 12 and $R_{\mathrm{miss}}$ to below 7\%. As shown in Fig.~\ref{fig:misregistration}, shifted infrared targets form separate bright contours beside the visible targets, interfering with recognition and localization by the detectors. In contrast, NatureFusion maintains $\mathrm{AP}_{50:95}$ between 32.68 and 36.21 under shifts of 8 to 32 pixels, close to the 35.84 of the original visible input, while still recovering about 22\% of pedestrians missed in visible images. This stems from NatureFusion using the visible input to constrain natural appearance while drawing complementary target cues from infrared. Even when infrared is misaligned, the fused output retains the main scene information from the visible input, avoiding the severe detection degradation caused by overlaying incorrect infrared contours.

\textbf{Teacher and student comparison.} On the LLVIP-NCE test split, we compare the teacher demonstrations used to construct the dataset, Fusion-only, and NatureFusion, with the original visible images as a reference. Teacher demonstrations are generated through an online image editing API using person bounding box annotations and infrared guidance. Both student configurations receive only complete visible and infrared images at inference. Detection metrics are averaged over the seven frozen detectors, and NIQE is computed on the same images.

Fig.~\ref{fig:teacher_student} shows person locations in the teacher and student outputs. In the red circular inset, the teacher demonstration incorrectly shifts the pedestrian to the right relative to the input. The pedestrian locations in Fusion-only and NatureFusion are closer to the original image. In this example, the students learn appearance from the demonstrations while using the inputs to preserve target layout, showing robustness to the teacher's positional error.

Table~\ref{tab:teacher_student} quantitatively compares teacher demonstrations with student outputs. Although the teacher additionally uses person boxes when generating demonstrations, Fusion-only and NatureFusion achieve $\mathrm{AP}_{50:95}$ scores 9.08 and 9.82 percentage points higher than the teacher, respectively, with further reductions in NIQE. As shown in Fig.~\ref{fig:teacher_student}, the teacher may shift target positions when generating demonstrations individually. By learning natural appearance from multiple demonstrations and using location and contour cues from the input pair, the student better preserves target layout and reduces such errors. The pretrained backbone's natural-image prior also helps improve the visual quality of the generated results. Task adaptation further increases $R_{\mathrm{miss}}$ from 29.0\% to 33.9\%, improving recognition of targets that are difficult to discern in visible images.

\subsection{Efficiency Analysis}
\label{subsec:efficiency}

\textbf{Generation time.} Fig.~\ref{fig:efficiency} compares generation time at native resolution for the seven multi-modal fusion methods and NatureFusion. We use ten fixed image pairs from each of MFNet, SMOD, M3FD, and LLVIP and repeat the measurements for three rounds after warm-up on the same RTX 3090. Timing starts with decoded inputs on the CPU and ends when the RGB output is available on the CPU. It includes preprocessing and necessary data and model transfers but excludes model loading and file I/O. NatureFusion processes a queue of ten images in stages with a batch size of 1. We report amortized time per image by dividing the total queue processing time by the number of images. Its times on the four sets are 7.81, 9.19, 16.71, and 29.09 seconds, respectively, corresponding to speedup factors of 1.6 to 2.7 over DuS-DiFuse and 5.2 to 9.3 over DiTFuse. It remains slower than the other five multi-modal fusion methods.

\textbf{Quality and efficiency.} Fig.~\ref{fig:efficiency_pareto}(a)--(d) compares performance and mean generation time across methods on four datasets. SMOD and M3FD use the perceptual quality metric MUSIQ and the fusion metric SD, respectively. SD reflects the dispersion of image intensities and image contrast. MFNet-250 and LLVIP-250 use performance averaged over seven frozen segmenters and detectors, respectively. NatureFusion achieves the best result on all four metrics. Compared with DuS-DiFuse and DiTFuse, it generates images faster while maintaining its lead in quality and downstream task performance. The other five fusion methods take less time but score lower on all four metrics. These results show that introducing a pretrained generative backbone incurs additional computation but allows NatureFusion to use its natural-image prior to produce higher-quality fused images and improve downstream task performance.

\section{Conclusion}
\label{sec:conclusion}

We reformulate infrared and visible image fusion as natural complementary enhancement based on the needs of human observers and downstream vision models. The central idea is to use infrared target locations and contours to supplement poorly observed regions in low-light visible images, then combine visible scene context with generative priors to render complementary information as natural visible appearance. NatureFusion connects natural appearance learning with improved target discernibility through offline teacher demonstrations, conditional flow matching, and decoupled task adaptation. The fusion adapter learns a natural fusion mapping from teacher demonstrations guided by both modalities. The task adapter further improves target presentation while keeping the parameters of this mapping fixed. Inference requires only a registered image pair and a fixed prompt. Experiments evaluate perceptual quality, recall of pedestrians missed on visible images, and detection and segmentation across multiple models. NatureFusion outperforms the compared multi-modal fusion and single-modality enhancement methods on all six no-reference perceptual metrics across three datasets and achieves better overall performance with multiple frozen vision models and retrained segmenters. Ablations further show that infrared guidance, fusion learning, and task adaptation contribute to supervision construction, natural appearance generation, and target discernibility, respectively. Together, these results support organizing cross-modal complementary information through natural appearance to serve both human perception and machine vision.

\bibliographystyle{IEEEtran}
\bibliography{main}

@article{mittal2013niqe,
  author = {Mittal, Anish and Soundararajan, Rajiv and Bovik, Alan C.},
  title = {Making a ``completely blind'' image quality analyzer},
  journal = {IEEE Signal Processing Letters},
  year = {2013},
  volume={20},
  number={3},
  pages={209--212},
}

@inproceedings{ho2021cfg,
  author       = {Ho, Jonathan and Salimans, Tim},
  title        = {Classifier-free diffusion guidance},
  booktitle    = {NeurIPS 2021 Workshop on Deep Generative Models and Downstream Applications},
  year         = {2021},
}

@article{jiang2023diffll,
  author       = {Jiang, Hai and Luo, Ao and Fan, Haoqiang and Han, Songchen and Liu, Shuaicheng},
  title        = {Low-light image enhancement with wavelet-based diffusion models},
  journal      = {ACM Transactions on Graphics},
  volume       = {42},
  number       = {6},
  pages        = {1--14},
  year         = {2023},
}

@inproceedings{wang2024quadprior,
  author       = {Wang, Wenjing and Yang, Huan and Fu, Jianlong and Liu, Jiaying},
  title        = {Zero-reference low-light enhancement via physical quadruple priors},
  booktitle    = {Proceedings of the IEEE/CVF Conference on Computer Vision and Pattern Recognition (CVPR)},
  year         = {2024},
  pages        = {26057--26066},
}

@inproceedings{feijoo2025darkir,
  author       = {Feijoo, Daniel and Benito, Juan C. and Garcia, Alvaro and Conde, Marcos V.},
  title        = {{DarkIR}: Robust low-light image restoration},
  booktitle    = {Proceedings of the IEEE/CVF Conference on Computer Vision and Pattern Recognition (CVPR)},
  year         = {2025},
  pages        = {10879--10889},
}

@article{yi2025diffretinexpp,
  author       = {Yi, Xunpeng and Xu, Han and Zhang, Hao and Tang, Linfeng and Ma, Jiayi},
  title        = {{Diff-Retinex++}: Retinex-driven reinforced diffusion model for low-light image enhancement},
  journal      = {IEEE Transactions on Pattern Analysis and Machine Intelligence},
  volume       = {47},
  number       = {8},
  pages        = {6823--6841},
  year         = {2025},
}

@inproceedings{cordts2016cityscapes,
  author = {Cordts, Marius and Omran, Mohamed and Ramos, Sebastian and Rehfeld, Timo and Enzweiler, Markus and Benenson, Rodrigo and Franke, Uwe and Roth, Stefan and Schiele, Bernt},
  title = {The {Cityscapes} dataset for semantic urban scene understanding},
  booktitle = {Proceedings of the IEEE Conference on Computer Vision and Pattern Recognition (CVPR)},
  year = {2016},
  pages = {3213--3223},
}

@article{zhang2020dbcnn,
  author = {Zhang, Weixia and Ma, Kede and Yan, Jia and Deng, Dexiang and Wang, Zhou},
  title = {Blind image quality assessment using a deep bilinear convolutional neural network},
  journal = {IEEE Transactions on Circuits and Systems for Video Technology},
  year = {2020},
  volume = {30},
  number = {1},
  pages = {36--47},
}

@inproceedings{ke2021musiq,
  author = {Ke, Junjie and Wang, Qifei and Wang, Yilin and Milanfar, Peyman and Yang, Feng},
  title = {{MUSIQ}: Multi-scale image quality transformer},
  booktitle = {Proceedings of the IEEE/CVF International Conference on Computer Vision (ICCV)},
  year = {2021},
  pages = {5128--5137},
}

@inproceedings{yang2022maniqa,
  author = {Yang, Sidi and Wu, Tianhe and Shi, Shuwei and Lao, Shanshan and Gong, Yuan and Cao, Mingdeng and Wang, Jiahao and Yang, Yujiu},
  title = {{MANIQA}: Multi-dimension attention network for no-reference image quality assessment},
  booktitle = {Proceedings of the IEEE/CVF Conference on Computer Vision and Pattern Recognition Workshops (CVPRW)},
  year = {2022},
  pages={1190--1199},
}

@inproceedings{agnolucci2024arniqa,
  author = {Agnolucci, Lorenzo and Galteri, Leonardo and Bertini, Marco and Del Bimbo, Alberto},
  title = {{ARNIQA}: Learning distortion manifold for image quality assessment},
  booktitle = {Proceedings of the IEEE/CVF Winter Conference on Applications of Computer Vision (WACV)},
  year = {2024},
  pages={188--197},
}

@inproceedings{wang2023clipiqa,
  author = {Wang, Jianyi and Chan, Kelvin C. K. and Loy, Chen Change},
  title = {Exploring {CLIP} for assessing the look and feel of images},
  booktitle = {Proceedings of the AAAI Conference on Artificial Intelligence},
  year = {2023},
  volume = {37},
  number = {2},
  pages = {2555--2563},
}

@inproceedings{zhang2023liqe,
  author = {Zhang, Weixia and Zhai, Guangtao and Wei, Ying and Yang, Xiaokang and Ma, Kede},
  title = {Blind image quality assessment via vision-language correspondence: A multitask learning perspective},
  booktitle = {Proceedings of the IEEE/CVF Conference on Computer Vision and Pattern Recognition (CVPR)},
  year = {2023},
  pages = {14071--14081},
}

@inproceedings{jia2021llvip,
  author       = {Jia, Xinyu and Zhu, Chuang and Li, Minzhen and Tang, Wenqi and Zhou, Wenli},
  title        = {{LLVIP}: A visible-infrared paired dataset for low-light vision},
  booktitle    = {Proceedings of the IEEE/CVF International Conference on Computer Vision Workshops (ICCVW)},
  year         = {2021},
  pages        = {3489--3497},
}

@article{tang2023divfusion,
  author       = {Tang, Linfeng and Xiang, Xinyu and Zhang, Hao and Gong, Meiqi and Ma, Jiayi},
  title        = {{DIVFusion}: Darkness-free infrared and visible image fusion},
  journal      = {Information Fusion},
  volume       = {91},
  pages        = {477--493},
  year         = {2023},
}

@inproceedings{liu2022tardal,
  author       = {Liu, Jinyuan and Fan, Xin and Huang, Zhanbo and Wu, Guanyao and Liu, Risheng and Zhong, Wei and Luo, Zhongxuan},
  title        = {Target-aware dual adversarial learning and a multi-scenario multi-modality benchmark to fuse infrared and visible for object detection},
  booktitle    = {Proceedings of the IEEE/CVF Conference on Computer Vision and Pattern Recognition (CVPR)},
  year         = {2022},
  pages        = {5792--5801},
}

@article{tang2026dspfusion,
  author       = {Tang, Linfeng and Li, Chunyu and Wang, Yeda and Wang, Guoqing and Yuan, Yixuan and Ma, Jiayi},
  title        = {{DSPFusion}: Image fusion via degradation and semantic dual-prior guidance},
  journal      = {IEEE Transactions on Image Processing},
  volume       = {35},
  pages        = {6331--6345},
  year         = {2026},
}

@inproceedings{cai2023retinexformer,
  author       = {Cai, Yuanhao and Bian, Hao and Lin, Jing and Wang, Haoqian and Timofte, Radu and Zhang, Yulun},
  title        = {{Retinexformer}: One-stage {Retinex}-based transformer for low-light image enhancement},
  booktitle    = {Proceedings of the IEEE/CVF International Conference on Computer Vision (ICCV)},
  year         = {2023},
  pages        = {12470--12479},
}

@inproceedings{zhao2023ddfm,
  author       = {Zhao, Zixiang and Bai, Haowen and Zhu, Yuanzhi and Zhang, Jiangshe and Xu, Shuang and Zhang, Yulun and Zhang, Kai and Meng, Deyu and Timofte, Radu and Van Gool, Luc},
  title        = {{DDFM}: Denoising diffusion model for multi-modality image fusion},
  booktitle    = {Proceedings of the IEEE/CVF International Conference on Computer Vision (ICCV)},
  year         = {2023},
  pages        = {8048--8059},
}

@inproceedings{cho2026mrilluminate,
  author       = {Cho, Joshua and Aghajanzadeh, Sara and Zhu, Zhen and Forsyth, David},
  title        = {{MR. Illuminate}: Zero-shot low-light image enhancement with diffusion prior},
  booktitle    = {Proceedings of the IEEE/CVF Conference on Computer Vision and Pattern Recognition (CVPR)},
  year         = {2026},
  pages        = {8460--8470},
}

@inproceedings{liu2025dcevo,
  author       = {Liu, Jinyuan and Zhang, Bowei and Mei, Qingyun and Li, Xingyuan and Zou, Yang and Jiang, Zhiying and Ma, Long and Liu, Risheng and Fan, Xin},
  title        = {{DCEvo}: Discriminative cross-dimensional evolutionary learning for infrared and visible image fusion},
  booktitle    = {Proceedings of the IEEE/CVF Conference on Computer Vision and Pattern Recognition (CVPR)},
  year         = {2025},
  pages        = {2226--2235},
}

@article{tang2022seafusion,
  author       = {Tang, Linfeng and Yuan, Jiteng and Ma, Jiayi},
  title        = {Image fusion in the loop of high-level vision tasks: A semantic-aware real-time infrared and visible image fusion network},
  journal      = {Information Fusion},
  volume       = {82},
  pages        = {28--42},
  year         = {2022},
}

@inproceedings{lipman2023flowmatching,
  author       = {Yaron Lipman and Ricky T. Q. Chen and Heli Ben-Hamu and Maximilian Nickel and Matthew Le},
  title        = {Flow matching for generative modeling},
  booktitle    = {Proceedings of the International Conference on Learning Representations},
  year         = {2023},
}

@article{tang2022piafusion,
  author       = {Tang, Linfeng and Yuan, Jiteng and Zhang, Hao and Jiang, Xingyu and Ma, Jiayi},
  title        = {{PIAFusion}: A progressive infrared and visible image fusion network based on illumination aware},
  journal      = {Information Fusion},
  volume       = {83--84},
  pages        = {79--92},
  year         = {2022},
}

@inproceedings{liu2023segmif,
  author       = {Liu, Jinyuan and Liu, Zhu and Wu, Guanyao and Ma, Long and Liu, Risheng and Zhong, Wei and Luo, Zhongxuan and Fan, Xin},
  title        = {Multi-interactive feature learning and a full-time multi-modality benchmark for image fusion and segmentation},
  booktitle    = {Proceedings of the IEEE/CVF International Conference on Computer Vision (ICCV)},
  year         = {2023},
  pages        = {8081--8090},
}

@inproceedings{bai2025tdfusion,
  author       = {Bai, Haowen and Zhang, Jiangshe and Zhao, Zixiang and Wu, Yichen and Deng, Lilun and Cui, Yukun and Feng, Tao and Xu, Shuang},
  title        = {Task-driven image fusion with learnable fusion loss},
  booktitle    = {Proceedings of the IEEE/CVF Conference on Computer Vision and Pattern Recognition (CVPR)},
  year         = {2025},
  pages        = {7457--7468},
}

@inproceedings{hu2022lora,
  author       = {Hu, Edward J. and Shen, Yelong and Wallis, Phillip and Allen-Zhu, Zeyuan and Li, Yuanzhi and Wang, Shean and Wang, Lu and Chen, Weizhu},
  title        = {{LoRA}: Low-rank adaptation of large language models},
  booktitle    = {Proceedings of the International Conference on Learning Representations (ICLR)},
  year         = {2022},
}

@inproceedings{ha2017mfnet,
  author       = {Ha, Qishen and Watanabe, Kohei and Karasawa, Takumi and Ushiku, Yoshitaka and Harada, Tatsuya},
  title        = {{MFNet}: Towards real-time semantic segmentation for autonomous vehicles with multi-spectral scenes},
  booktitle    = {Proceedings of the IEEE/RSJ International Conference on Intelligent Robots and Systems (IROS)},
  year         = {2017},
  pages        = {5108--5115},
}

@inproceedings{yan2025hvi,
  author       = {Yan, Qingsen and Feng, Yixu and Zhang, Cheng and Pang, Guansong and Shi, Kangbiao and Wu, Peng and Dong, Wei and Sun, Jinqiu and Zhang, Yanning},
  title        = {{HVI}: A new color space for low-light image enhancement},
  booktitle    = {Proceedings of the IEEE/CVF Conference on Computer Vision and Pattern Recognition (CVPR)},
  year         = {2025},
  pages        = {5678--5687},
}

@inproceedings{zhang2026magicfuse,
  author       = {Zhang, Hao and Zha, Yanping and Li, Zizhuo and Gong, Meiqi and Ma, Jiayi},
  title        = {{MagicFuse}: Single image fusion for visual and semantic reinforcement},
  booktitle    = {Proceedings of the IEEE/CVF Conference on Computer Vision and Pattern Recognition (CVPR)},
  year         = {2026},
  pages        = {26551--26560},
}

@inproceedings{yi2024textif,
  author       = {Yi, Xunpeng and Xu, Han and Zhang, Hao and Tang, Linfeng and Ma, Jiayi},
  title        = {{Text-IF}: Leveraging semantic text guidance for degradation-aware and interactive image fusion},
  booktitle    = {Proceedings of the IEEE/CVF Conference on Computer Vision and Pattern Recognition (CVPR)},
  year         = {2024},
  pages={27016--27025},
}

@article{zhang2025omnifuse,
  author       = {Zhang, Hao and Cao, Lei and Zuo, Xuhui and Shao, Zhenfeng and Ma, Jiayi},
  title        = {{OmniFuse}: Composite degradation-robust image fusion with language-driven semantics},
  journal      = {IEEE Transactions on Pattern Analysis and Machine Intelligence},
  year         = {2025},
  volume       = {47},
  number       = {9},
  pages        = {7577--7595},
}

@article{cao2026dusdifuse,
  author       = {Cao, Lei and Zhang, Hao and Zhang, Peng and Zhou, Daiguo and Ma, Jiayi},
  title        = {Towards heterogeneous-degradation-robust image fusion with controllable generative modulation},
  journal      = {IEEE Transactions on Pattern Analysis and Machine Intelligence},
  year         = {2026},
  pages        = {1--18},
  note         = {Early access},
}

@article{li2026ditfuse,
  author       = {Li, Jiayang and Jiang, Chengjie and Jiang, Junjun and Liang, Pengwei and Ma, Jiayi and Nie, Liqiang},
  title        = {Towards unified semantic and controllable image fusion: A diffusion transformer approach},
  journal      = {IEEE Transactions on Pattern Analysis and Machine Intelligence},
  year         = {2026},
  volume       = {48},
  number       = {4},
  pages        = {3970--3987},
}

@inproceedings{guo2025hclfuse,
  author       = {Guo, Lin and Luo, Xiaoqing and Xie, Wei and Zhang, Zhancheng and Li, Hui and Wang, Rui and Feng, Zhenhua and Song, Xiaoning},
  title        = {Revisiting generative infrared and visible image fusion based on human cognitive laws},
  booktitle    = {Advances in Neural Information Processing Systems},
  volume={38},
  pages={96322--96352},
  year={2025}
}

@inproceedings{wang2026isfuse,
  author       = {Wang, Xue and Guan, Zheng and Qian, Wenhua and Wang, Chengchao and Ma, Runzhuo},
  title        = {Multi-modal image fusion via intervention-stable feature learning},
  booktitle    = {Proceedings of the IEEE/CVF Conference on Computer Vision and Pattern Recognition (CVPR)},
  year         = {2026},
  pages        = {33827--33837},
}

@article{chen2025amfd,
  author = {Chen, Zizhao and Qian, Yeqiang and Yang, Xiaoxiao and Wang, Chunxiang and Yang, Ming},
  title = {{AMFD}: Distillation via adaptive multimodal fusion for multispectral pedestrian detection},
  journal = {IEEE Transactions on Multimedia},
  year = {2025},
  volume = {27},
  pages = {8298--8310},
}

@misc{bfl2026flux2klein,
  author       = {{Black Forest Labs}},
  title        = {{FLUX.2 [klein]}: Towards interactive visual intelligence},
  year         = {2026},
  month        = jan,
  howpublished = {Black Forest Labs blog},
}

@inproceedings{lin2014coco,
  author       = {Lin, Tsung-Yi and Maire, Michael and Belongie, Serge and Hays, James and Perona, Pietro and Ramanan, Deva and Doll\'{a}r, Piotr and Zitnick, C. Lawrence},
  title        = {{Microsoft COCO}: Common objects in context},
  year         = {2014},
  booktitle    = {Proceedings of the European Conference on Computer Vision},
  pages        = {740--755},
}

@misc{jocher2024yolo11,
  author       = {Jocher, Glenn and Qiu, Jing},
  title        = {{Ultralytics YOLO11}},
  year         = {2024},
  howpublished = {Official Software and Model Documentation},
}

@misc{ultralytics2026yolo26,
  author       = {Jocher, Glenn and Qiu, Jing and Liu, Mengyu and Lyu, Shuai and Akyon, Fatih Cagatay and Kalfaoglu, Muhammet Esat},
  title        = {{Ultralytics YOLO26}: Unified real-time end-to-end vision models},
  year         = {2026},
  howpublished = {{arXiv} preprint arXiv:2606.03748},
}

@inproceedings{ren2015fasterrcnn,
  author       = {Ren, Shaoqing and He, Kaiming and Girshick, Ross and Sun, Jian},
  title        = {{Faster R-CNN}: Towards real-time object detection with region proposal networks},
  year         = {2015},
  booktitle    = {Advances in Neural Information Processing Systems},
  volume       = {28},
}

@inproceedings{lin2017focalloss,
  author       = {Lin, Tsung-Yi and Goyal, Priya and Girshick, Ross and He, Kaiming and Doll\'{a}r, Piotr},
  title        = {Focal loss for dense object detection},
  year         = {2017},
  booktitle    = {Proceedings of the IEEE International Conference on Computer Vision (ICCV)},
  pages        = {2980--2988},
}

@misc{lv2024rtdetrv2,
  author       = {Lv, Wenyu and Zhao, Yian and Chang, Qinyao and Huang, Kui and Wang, Guanzhong and Liu, Yi},
  title        = {{RT-DETRv2}: Improved baseline with bag-of-freebies for real-time detection transformer},
  year         = {2024},
  howpublished = {{arXiv} preprint arXiv:2407.17140},
}

@inproceedings{peng2025dfine,
  author       = {Peng, Yansong and Li, Hebei and Wu, Peixi and Zhang, Yueyi and Sun, Xiaoyan and Wu, Feng},
  title        = {{D-FINE}: Redefine regression task of {DETRs} as fine-grained distribution refinement},
  year         = {2025},
  booktitle    = {Proceedings of the International Conference on Learning Representations (ICLR)},
  pages={44015--44031},
  volume={2025},
}

@inproceedings{zhao2017pspnet,
  author       = {Zhao, Hengshuang and Shi, Jianping and Qi, Xiaojuan and Wang, Xiaogang and Jia, Jiaya},
  title        = {Pyramid scene parsing network},
  year         = {2017},
  booktitle    = {Proceedings of the IEEE Conference on Computer Vision and Pattern Recognition (CVPR)},
  pages        = {2881--2890},
}

@inproceedings{chen2018deeplabv3plus,
  author       = {Chen, Liang-Chieh and Zhu, Yukun and Papandreou, George and Schroff, Florian and Adam, Hartwig},
  title        = {Encoder-decoder with atrous separable convolution for semantic image segmentation},
  year         = {2018},
  booktitle    = {Proceedings of the European Conference on Computer Vision (ECCV)},
  pages={833--851},
}

@inproceedings{yuan2020ocrnet,
  author       = {Yuan, Yuhui and Chen, Xilin and Wang, Jingdong},
  title        = {Object-contextual representations for semantic segmentation},
  year         = {2020},
  booktitle    = {Proceedings of the European Conference on Computer Vision},
  pages        = {173--190},
}

@inproceedings{xie2021segformer,
  author       = {Xie, Enze and Wang, Wenhai and Yu, Zhiding and Anandkumar, Anima and Alvarez, Jose M. and Luo, Ping},
  title        = {{SegFormer}: Simple and efficient design for semantic segmentation with transformers},
  year         = {2021},
  booktitle    = {Advances in Neural Information Processing Systems},
  volume={34},
  pages={12077--12090},
}

@inproceedings{cheng2022mask2former,
  author       = {Cheng, Bowen and Misra, Ishan and Schwing, Alexander G. and Kirillov, Alexander and Girdhar, Rohit},
  title        = {Masked-attention mask transformer for universal image segmentation},
  year         = {2022},
  booktitle    = {Proceedings of the IEEE/CVF Conference on Computer Vision and Pattern Recognition (CVPR)},
  pages        = {1280--1289},
}

@inproceedings{fu2025segman,
  author       = {Fu, Yunxiang and Lou, Meng and Yu, Yizhou},
  title        = {{SegMAN}: Omni-scale context modeling with state space models and local attention for semantic segmentation},
  year         = {2025},
  booktitle    = {Proceedings of the IEEE/CVF Conference on Computer Vision and Pattern Recognition (CVPR)},
  pages        = {19077--19087},
}

@inproceedings{cai2018cascadercnn,
  author = {Cai, Zhaowei and Vasconcelos, Nuno},
  title = {{Cascade R-CNN}: Delving into high quality object detection},
  booktitle = {Proceedings of the IEEE/CVF Conference on Computer Vision and Pattern Recognition (CVPR)},
  year = {2018},
  pages = {6154--6162}
}

@article{hosu2020koniq,
  author       = {Hosu, Vlad and Lin, Hanhe and Szir\'{a}nyi, Tam\'{a}s and Saupe, Dietmar},
  title        = {{KonIQ-10k}: An ecologically valid database for deep learning of blind image quality assessment},
  year         = {2020},
  journal      = {IEEE Transactions on Image Processing},
  volume       = {29},
  pages        = {4041--4056},
}

@misc{chen2022pyiqa,
  author       = {Chen, Chaofeng and Mo, Jiadi},
  title        = {{IQA-PyTorch}: {PyTorch} toolbox for image quality assessment},
  year         = {2022},
  howpublished = {Open-Source Software},
}

@article{yi2026ll,
  title={LL-Refiner: Learning Adaptive Refinement for Ultra-High-Definition Low-Light Image Enhancement},
  author={Yi, Xunpeng and Yan, Qinglong and Zhang, Yibing and Li, Zongrong and Zhang, Hao and Cai, Zhanchuan and Ma, Jiayi},
  journal={IEEE/CAA Journal of Automatica Sinica},
  volume={13},
  number={6},
  pages={1288--1300},
  year={2026}
}

\end{document}